\documentclass{article}

\PassOptionsToPackage{numbers,sort&compress}{natbib}

\usepackage[final,eandd]{neurips_2026}

\usepackage[utf8]{inputenc} 
\usepackage[T1]{fontenc}    
\usepackage{hyperref}       
\usepackage{url}            
\usepackage{booktabs}       
\usepackage{amsfonts}       
\usepackage{nicefrac}       
\usepackage{microtype}      
\usepackage{xcolor}         
\usepackage{longtable}

\usepackage{amsmath}
\usepackage{amsthm}
\usepackage{amssymb}
\usepackage{multirow}
\usepackage{graphicx}
\usepackage{array}
\usepackage{tabularx}
\usepackage{enumitem}

\usepackage{tikz}
\usepackage{placeins}

\usetikzlibrary{arrows.meta, positioning, calc}
\definecolor{dtblue}{HTML}{3B6EA5}
\definecolor{dtbluefill}{HTML}{EDF2F8}
\definecolor{dtorange}{HTML}{D1832F}
\definecolor{dtorangefill}{HTML}{FBF1E5}
\definecolor{dtgray}{HTML}{8A8A8A}
\definecolor{dtink}{HTML}{333333}
\newcommand{\axsub}[1]{{\scriptsize\color{black!65}\strut #1}}

\newtheorem{proposition}{Proposition}

\newcommand{\libname}{\textsc{DTCD}}
\newcommand{\bN}{\mathbf{N}}

\title{Composable Causality: A Toolkit for Systematic Time-Series Causal Discovery and Treatment-Effect Benchmarking}

\author{%
  Muhammad Hasan Ferdous \\
  Causal AI Lab\\
  Department of Information Systems\\
  University of Maryland, Baltimore County\\
  \texttt{h.ferdous@umbc.edu}
  \And
  Md Osman Gani \\
  Causal AI Lab\\
  Department of Information Systems\\
  University of Maryland, Baltimore County\\
  \texttt{mogani@umbc.edu}
}

\hypersetup{
    pdftitle={Composable Causality: A Toolkit for Systematic Time-Series Causal Discovery and Treatment-Effect Benchmarking},
    pdfauthor={Muhammad Hasan Ferdous and Md Osman Gani},
    pdfsubject={NeurIPS 2026, Track on Evaluations and Datasets},
    pdfkeywords={causal discovery, time series, benchmark, treatment effect estimation}
}

\begin{document}

\maketitle

\begin{abstract}
Real-world multivariate time series rarely come with verified ground-truth causal structure, so research on time-series causal discovery and treatment-effect estimation depends on synthetic data. Existing tools are either fixed datasets that cannot be modified after release or script-based generators in which each script fixes one combination of mechanism, confounding, and missingness. Neither supports controlled experiments that vary a single property, such as sparsity, noise distribution, or missingness mechanism, while holding the others fixed. Counterfactual evaluation has the same gap, since existing simulators are built per paper with hardcoded mechanisms or produce observational data only. We introduce \libname{}\footnote{Code and data: \url{https://github.com/hferdous/Composable_Causality}.} (Datasets and Toolkit for Causal Discovery), a data generation library that produces multivariate time series with ground-truth contemporaneous and lagged adjacency matrices. Topology, mechanism, noise, observational mask, and intervention policy are independent components that compose through configuration alone. Paired counterfactual twins, produced by caching the stochastic realization during the factual rollout and re-injecting it during the interventional rollout, recover individual treatment effects to numerical precision. A reference benchmark of eight discovery algorithms on eighty-one configurations and four treatment-effect estimators on eighteen configurations exposes failure modes that aggregate metrics hide. These include the inability of standard discovery methods to handle missing values, regardless of whether the missingness mechanism is informative, and a marginal-variance shortcut (varsortability) that we measure and remove by standardization.
\end{abstract}

\section{Introduction}
\label{sec:intro}

The discovery of causal relationships from observational multivariate time 
series is a foundational prerequisite for understanding complex dynamic 
systems in disciplines spanning healthcare~\citep{ferdous2023cdans}, 
algorithmic finance, and climate 
science~\citep{hossain2025correlation,hossain2025causal}. Over the past decade, researchers have developed numerous 
algorithms for time-series causal discovery, including Granger-style 
methods, constraint-based independence testing, score-based continuous 
optimization, and deep functional approaches.

Despite this methodological proliferation, empirical evaluation has lagged. 
Existing benchmarks fall into two architectural patterns. The first is the 
\emph{static corpus}: a fixed collection of pre-generated datasets shipped 
alongside a paper or maintained as a community 
repository~\citep{runge2019inferring}. The second is the \emph{scenario 
script collection}: a set of generator programs, each hardcoding a specific 
combination of mechanism, confounding pattern, and missingness 
assumption~\citep{Ferdous2025TimeGraph}. Both patterns share a common 
limitation: researchers cannot ablate difficulty axes or run controlled 
sensitivity sweeps without authoring new generator code per configuration.

We argue this is the wrong abstraction for methodological research. The 
questions that drive method development (\emph{at what sparsity does 
constraint-based discovery break? how does method ranking change as 
nonlinearity intensifies? what is the marginal cost of MCAR vs MNAR 
under otherwise identical conditions?}) require factorial experiments in 
which one data-generating process (DGP) axis varies while all others are 
held fixed. Static corpora and per-scenario scripts do not support this 
workflow.

\textbf{Contributions.} We introduce \libname{}, a configurable framework 
for temporal benchmarking that replaces the fixed-artifact paradigm.

\begin{enumerate}
    \item \textbf{A composable framework rather than a corpus or script 
    collection.} Every DGP component (topology generator, structural 
    mechanism, noise distribution, observational mask, intervention policy) 
    is a first-class API parameter that composes orthogonally with every 
    other component. Investigations that require new code in script-based 
    benchmarks (sensitivity sweeps, factorial ablations, controlled 
    perturbation studies) require only configuration changes in \libname{}.

    \item \textbf{Exact counterfactual twin generation.} The 
    \texttt{CounterfactualSuite} caches the stochastic realization $\bN$ 
    during the factual rollout and re-injects it during the interventional 
    rollout, so the individual treatment effect at any time $t$ for any 
    downstream node is exact to floating-point precision. No prior temporal 
    benchmark provides ground-truth counterfactuals.

    \item \textbf{Explicit identifiability documentation.} Identifiability status 
    (full DAG, Markov equivalent, or not established) is documented for the 
    linear mechanisms, as propositions, and for the recursive nonlinear 
    mechanism, which is used only as a robustness stressor. No claim is 
    made for the remaining mechanisms. 
    Machine-readable tags currently exist only in the spatio-temporal 
    module; adding them to the core generators is planned.

    \item \textbf{A reference benchmark and diagnostic protocol.} We 
    evaluate eight causal discovery algorithms and four CATE estimators 
    across a tractable factorial grid of nine predefined graphs 
    ($d \in \{4,6,8\}, \tau_{\max} \in \{2,3,4\}$) under nine discovery 
    DGP variants (linear, nonlinear, drift, MNAR, MCAR, trend, seasonal, 
    trend+seasonal, nonlinear+trend+seasonal) and two CATE variants. 
    Tuning uses Optuna with standardized metrics (SHD, normalized SHD, F1, 
    TPR, FDR, AUPRC, PEHE), and recovered graphs are visualized beside 
    ground truth to surface failure modes aggregate metrics conceal.

    \item \textbf{Open-source release with reproducibility infrastructure.} 
    Native exporters to \texttt{tigramite} and \texttt{causal-learn}, 
    Datasheet for Datasets~\citep{gebru2021datasheets} documentation, 
    Croissant~\citep{croissant} metadata, single-command reproduction 
    of every paper table and figure.
\end{enumerate}

\section{Related work}
\label{sec:related}

\paragraph{Time-series causal discovery methods.}
Time-series causal discovery comprises four broad paradigms. Granger-style
methods extend classical linear prediction tests~\citep{granger1969investigating},
and VAR-LiNGAM~\citep{hyvarinen2010estimation} provides identifiability
under non-Gaussian noise. Constraint-based methods evaluate conditional
independence on lagged variables. PCMCI~\citep{runge2019detecting} handles
high-dimensional autocorrelated data and PCMCI+~\citep{runge2020discovering}
adds contemporaneous links, while LPCMCI~\citep{gerhardus2020high} relaxes
causal sufficiency and CD-NOD~\citep{huang2020causal} exploits
non-stationarity as an identifiability resource. Recent work targets
non-stationary regimes directly. FANTOM~\citep{rahmani2025fantom} jointly
infers unknown regimes and a DAG per regime under non-Gaussian or
heteroscedastic noise, CDANs~\citep{ferdous2023cdans,ferdous2023ecdans} identifies lagged and
contemporaneous edges together with the variables whose mechanisms change
over time, DCD~\citep{ferdous2026dcd} targets autocorrelated and
non-stationary data through a decomposition-based approach, and mixture causal discovery~\citep{varambally2024discovering}
recovers one graph per latent component when each series comes from one of
several structural causal models. \libname{} can emulate these settings by
composition (parent-dependent or drifting noise scales, scheduled switches
between adjacency structures, random-walk edge weights, per-series mixtures
of configurations); benchmarking methods built for them is left to future
work. Score-based continuous optimization frameworks adapted from
NOTEARS~\citep{zheng2018notears} include DYNOTEARS~\citep{pamfil2020dynotears},
the nonparametric NTS-NOTEARS~\citep{sun2023ntsnotears}, and the deep
flow-based Rhino~\citep{gong2022rhino}. Neural functional approaches include
attention-based TCDF~\citep{nauta2019causal}, amortized causal
discovery~\citep{lowe2022amortized}, and component-wise MLP and LSTM models
for nonlinear Granger causality~\citep{tank2022neural}.
CUTS+~\citep{cheng2024cuts} addresses high-dimensional time series with
missing data via joint neural imputation and discovery.

\paragraph{Static benchmarks and dataset corpora.}
The CauseMe platform~\citep{runge2019inferring} hosts time series with known
ground truth from synthetic models that mimic real-data challenges and serves
as a common reference for comparing methods. The DREAM
challenges~\citep{prill2010dream3,marbach2009dream}, designed for
gene-regulatory network inference, are often repurposed as time-series
benchmarks despite their biology-specific generative process.
NetSim~\citep{smith2011network} provides fixed simulated datasets for
evaluating fMRI connectivity methods. Each of these resources fixes the data
at release, so researchers cannot ablate difficulty axes without writing new
generation code.

\paragraph{Script-based and configurable benchmark generators.}
TimeGraph~\citep{Ferdous2025TimeGraph} expanded coverage with
scenario-specific generators, each pairing a linear, polynomial, or
trigonometric mechanism with fixed choices of trend, seasonality, noise,
sampling, missingness, and confounding, so combining one scenario's mechanism
with another's noise or missingness requires modifying generator code.
\citet{lawrence2021data} propose a flexible data generating process that lets
users violate common assumptions and measure how methods degrade.
SynTReN~\citep{vandenbulcke2006syntren} simulates steady-state gene
expression data with hardcoded mechanisms. The \texttt{tigramite}
package~\citep{runge2019detecting} includes a parameterized generator of
structural causal processes and \texttt{causal-learn}~\citep{causallearn2024}
provides discovery algorithms and evaluation utilities, but neither composes
generation, masking, intervention, and evaluation into benchmark suites.
\citet{peters2013causal} formalized the distinction between full time graphs
and summary graphs that time-series benchmarks, including ours, inherit.
\libname{} keeps the full time graph, replaces per-scenario scripts with one
compositional API that also covers observation masks, interventions, and
exact counterfactual twins, and exports to \texttt{tigramite} and
\texttt{causal-learn} formats.

\paragraph{Treatment-effect estimation from time series.}
Synthetic control~\citep{abadie2010synthetic} estimates the effect of an
intervention on one treated unit from a weighted combination of untreated
units, relaxing the parallel trends assumption. Recurrent marginal structural
networks~\citep{lim2018forecasting} use recurrent networks to estimate
inverse probability of treatment weights for time-dependent confounding, and
the Counterfactual Recurrent Network (CRN)~\citep{bica2020crn} instead learns
balanced representations through adversarial training.
Representation-learning CATE estimators for static data include
TARNet~\citep{shalit2017estimating}, SNet~\citep{curth2021nonparametric}, and
FlexTENet~\citep{curth2021inductive}, while G-Net~\citep{li2021gnet} and the
Causal Transformer~\citep{melnychuk2022causal} estimate counterfactual
outcomes under time-varying treatments. Evaluation in this literature has
typically relied on per-paper simulators or on conditional expectations
rather than exact individual treatment effects. \libname{} supports a
standardized PEHE on exact counterfactual twins, which requires cached-noise
generation. CRN, G-Net, and the Causal Transformer are not yet in our
reference CATE benchmark because each needs a wrapper exposing a unified
\texttt{fit(X, A, mask) -> tau\_hat} interface; this is planned follow-up
work.

\paragraph{Concurrent work on temporal interventional and counterfactual data.}
Two recent works address adjacent but distinct problems. Thumm 
and Chen~\citep{thumm2026interventional} produce paired 
observational and interventional time series sampled from temporal 
structural causal models, with the goal of training prior-data 
fitted networks for in-context causal effect estimation. Their 
generated data serves as a training distribution for foundation 
models, not as a benchmark for evaluating discovery or estimation 
methods; the framework provides interventional rollouts but does 
not document identifiability status per mechanism, ship 
reference benchmark scripts, or expose discovery-method exporters. 
Xia et al.~\citep{xia2025causal} introduce a diffusion-based
generative model that produces interventional time series via backdoor
adjustment and counterfactual time series via abduction, action, and
prediction, so its counterfactuals come from a learned model rather than a
known structural equation. Their counterfactual quality is 
therefore subject to model fidelity, whereas \libname{}'s 
counterfactual twins are exact by construction. \libname{} is 
complementary to both: where \citet{thumm2026interventional} provide a training 
distribution and \citet{xia2025causal} provide a generative model, \libname{} 
provides a benchmark library with deterministic ground truth and 
documented identifiability status.

\section{Framework design}
\label{sec:framework}

\subsection{Data-generating process}

A temporal causal model over $d$ variables evolves according to structural 
equations:
\begin{equation}
X_{t,i} = f_i\left(\text{pa}(X_{t,i}), N_{t,i}\right),
\end{equation}
where $\text{pa}(X_{t,i})$ comprises both contemporaneous parents 
$X_{t,j} : (j \to i) \in \mathcal{A}_0$ and lagged parents 
$X_{t-\tau,j} : (j \to i, \tau) \in \mathcal{A}_\tau$ for $\tau \in 
\{1,\ldots,\tau_{\max}\}$, with $\tau_{\max}$ the maximum admissible lag for 
dependencies. The stochastic noise $N_{t,i}$ is independent across $i$ 
(causal sufficiency assumption) but may be autocorrelated within $i$ 
(supported via AR noise generators).

\begin{figure}[t]
\centering
\resizebox{0.84\textwidth}{!}{%
\begin{tikzpicture}[
  font=\small,
  axis/.style={draw=dtblue, line width=0.7pt, rounded corners=2.5pt,
               fill=dtbluefill, minimum width=2.52cm, minimum height=1.30cm,
               align=center, inner sep=2pt},
  outbox/.style={draw=dtorange, line width=0.7pt, rounded corners=2.5pt,
              fill=dtorangefill, minimum width=3.231cm, minimum height=0.98cm,
              align=center, inner sep=2pt},
  flow/.style={-{Latex[length=1.9mm]}, line width=0.6pt, dtgray}
]

\node[axis] (ax0) at (1.260, 0)
  {{\bfseries\color{dtblue}\strut Topology}\\[1pt] \axsub{ER / BA / WS}\\ \axsub{templates}};
\node[axis] (ax1) at (4.105, 0)
  {{\bfseries\color{dtblue}\strut Mechanism}\\[1pt] \axsub{linear / poly / MLP}\\ \axsub{Fourier / PNL}};
\node[axis] (ax2) at (6.950, 0)
  {{\bfseries\color{dtblue}\strut Noise}\\[1pt] \axsub{Gauss / Laplace / $t$}\\ \axsub{Cauchy / drift}};
\node[axis] (ax3) at (9.795, 0)
  {{\bfseries\color{dtblue}\strut Mask}\\[1pt] \axsub{full / MCAR}\\ \axsub{MNAR}};
\node[axis] (ax4) at (12.640, 0)
  {{\bfseries\color{dtblue}\strut Intervention}\\[1pt] \axsub{none / do}\\ \axsub{soft / policy}};

\node[draw=dtink, line width=0.8pt, rounded corners=2.5pt, fill=black!4,
      minimum width=13.9cm, minimum height=0.82cm, align=center]
  (roll) at (6.95, -1.78)
  {\strut Structural rollout\quad
   $X_{t,i} = f_i\bigl(\mathrm{pa}(X_{t,i}),\, N_{t,i}\bigr)$
   \quad {\footnotesize(noise realization $\mathbf{N}$ cached)}};

\foreach \i in {0,...,4} {\draw[flow] (ax\i.south) -- (ax\i.south |- roll.north);}

\node[outbox] (o0) at (1.616, -3.36)
  {\strut Trajectory\\[-1pt] {\scriptsize\strut $X \in \mathbb{R}^{T \times d}$}};
\node[outbox] (o1) at (5.172, -3.36)
  {\strut Ground truth\\[-1pt] {\scriptsize\strut $\{A_\tau\}_{\tau=0}^{\tau_{\max}}$}};
\node[outbox] (o2) at (8.728, -3.36)
  {\strut Observation mask\\[-1pt] {\scriptsize\strut $R \in \{0,1\}^{T \times d}$}};
\node[outbox] (o3) at (12.284, -3.36)
  {\strut Counterfactual twin\\[-1pt] {\scriptsize\strut (re-inject $\mathbf{N}$)}};

\foreach \i in {0,...,3} {\draw[flow] (o\i.north |- roll.south) -- (o\i.north);}

\end{tikzpicture}}
\caption{Compositional structure of \libname{}. Five independent configuration
axes feed one structural rollout, which emits the trajectory, the ground-truth
adjacencies, the observation mask, and, when an intervention policy is supplied,
an exactly paired counterfactual twin via re-injection of the cached noise
$\mathbf{N}$. Varying one axis holds all others fixed by construction.}
\label{fig:architecture}
\end{figure}
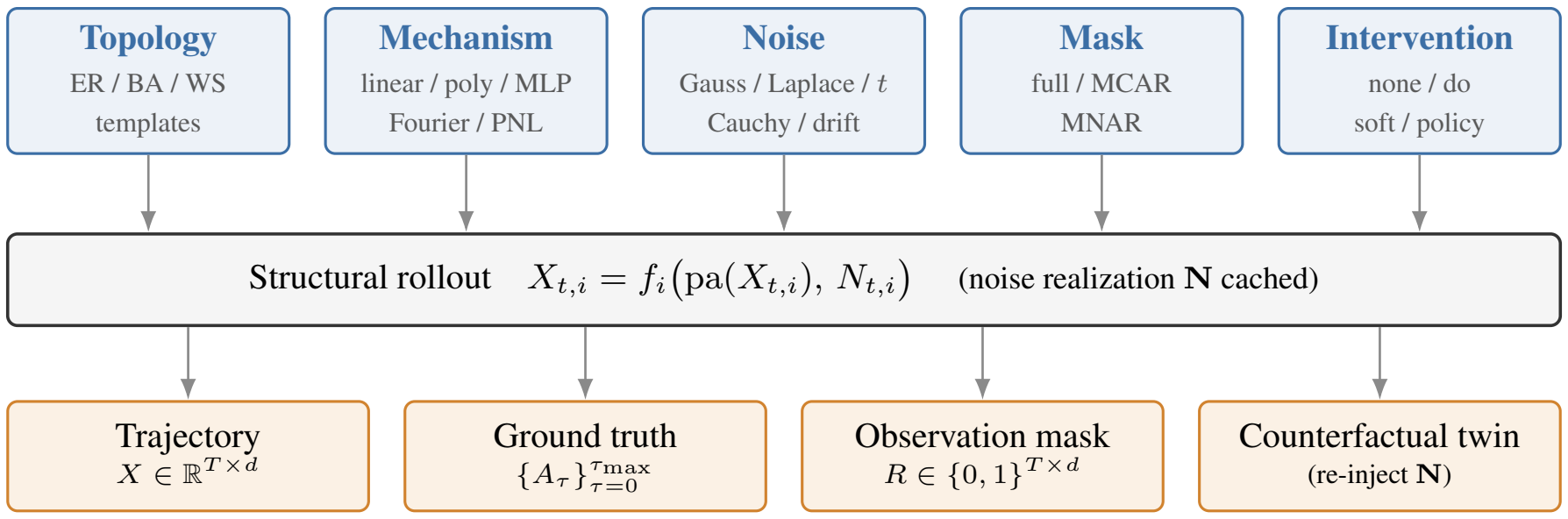

\libname{} decomposes DGP specification into orthogonal configuration axes
(Figure~\ref{fig:architecture}):

\paragraph{Graph topology.}
Topology generators accept $d, \tau_{\max}, \text{sparsity} \in [0,1]$ and 
return a binary adjacency tensor $\mathcal{A} \in \{0,1\}^{d \times d \times (\tau_{\max}+1)}$, 
where $\mathcal{A}_{i,j,\tau}=1$ indicates $X_{t-\tau,j} \to X_{t,i}$. 
Supported generators include Erd\H{o}s-R\'enyi, scale-free
Barab\'asi-Albert, and small-world Watts-Strogatz graphs, plus the nine
curated templates of the reference benchmark (Appendix~\ref{app:dgp-params}).

\paragraph{Structural mechanism.}
The functional form $f_i$ is linear, polynomial, tanh,
pairwise-interaction, Fourier-basis, or a multilayer perceptron, optionally
wrapped in a strictly monotone post-nonlinear distortion or combined
multiplicatively with the noise; softmax and Poisson mechanisms cover
discrete variables. Section~\ref{sec:identifiability} documents the identifiability
status of the linear mechanisms and of the recursive nonlinear one; no
identifiability claim is made for the other mechanisms. Linear Gaussian mechanisms identify only up to Markov equivalence; 
linear non-Gaussian mechanisms (VAR-LiNGAM-compatible) identify the full 
DAG; the nonlinear power-transform mechanism is not provably identifiable 
in its recursive form and is provided as a robustness stressor rather than 
an identifiability target.

\paragraph{Noise distribution.}
Supported noise models are Gaussian, Laplace, Student-$t$, exponential,
uniform, Cauchy, heteroscedastic, drifting-scale, and mixtures.  
Non-Gaussian noise is required for VAR-LiNGAM identifiability. Noise 
realizations are cached per timestep and variable during generation, 
enabling exact counterfactual queries (Section~\ref{sec:counterfactuals}).

\paragraph{Observational mask.}
Missing data patterns follow MCAR and MNAR mechanisms. The mask 
$R \in \{0,1\}^{T \times d}$ is generated jointly with $X$. Under MCAR, 
each entry is masked with probability $p$ independent of the value; under 
MNAR, masking is implemented via self-thresholding (each variable's 
top-quantile values are masked). Including both mechanisms enables 
controlled isolation of missing-value robustness from informativeness 
of the missingness pattern.

\paragraph{Exogenous components.}
Optional additive trend (linear in time) and seasonal (sum of sinusoids 
at configurable periods, supporting both single and multi-period 
superpositions such as daily $\times$ weekly $\times$ annual cycles) 
components can be applied per variable as post-hoc additions to the 
structural output:
\begin{equation*}
X^{\text{obs}}_{t,i} = X_{t,i} + \alpha_i \cdot t/T + \sum_k \beta_{i,k} \sin(2\pi t / P_k + \phi_{i,k}).
\end{equation*} These do not enter the structural 
equations and the ground-truth graph is unchanged across variants with 
or without them. Their inclusion probes method robustness to 
non-stationary observational artifacts that are common in real-world 
deployments but absent from canonical synthetic benchmarks.

\paragraph{Intervention policy.}
The CounterfactualSuite accepts $\text{do}(X_{\tau,i} = c)$ interventions 
at specified timesteps. Downstream effects propagate through the graph via 
the cached noise $\bN$. This construction yields exact individual treatment 
effects: $\text{ITE}_{t,j} = X^{\text{do}}_{t,j} - X^{\text{obs}}_{t,j}$ 
is floating-point exact because the only difference between factual and 
counterfactual rollouts is the intervened node value at the intervention 
timestep.

\subsection{Identifiability guarantees}
\label{sec:identifiability}

We document the identifiability class of the linear mechanisms as follows:

\begin{proposition}[Linear Gaussian VAR identifiability]
\label{prop:lingauss}
Suppose the data are generated by a stationary linear vector 
autoregression with: (i) causal sufficiency (no latent confounders), 
(ii) Gaussian innovations $N_{t,i} \sim \mathcal{N}(0, \sigma_i^2)$ 
with $\sigma_i > 0$, (iii) the contemporaneous adjacency matrix $A_0$ 
acyclic, and (iv) faithfulness (Markov properties imply structural 
ones). Then the supports of the lagged adjacency matrices
$\{A_\tau\}_{\tau=1}^{\tau_{\max}}$ are identified, while $A_0$ is
identifiable only up to the Markov equivalence class of the full time graph,
where lagged v-structures can orient some contemporaneous edges. This
follows from reduced-form VAR identifiability \citep{lutkepohl2005new} and
the graphical characterization of Markov
equivalence~\citep{verma1990equivalence,spirtes2000causation}.
\end{proposition}

\begin{proposition}[VAR-LiNGAM identifiability]
\label{prop:varlingam}
Suppose the data are generated by a stationary linear vector 
autoregression with: (i) causal sufficiency, (ii) non-Gaussian, 
mutually independent innovations $N_{t,i}$ with at most one 
Gaussian-distributed component, (iii) acyclicity at each time slice, 
and (iv) the contemporaneous adjacency matrix $A_0$ can be permuted to a
strictly lower-triangular matrix (LiNGAM acyclicity condition).
Then both $A_0$ and $\{A_\tau\}_{\tau=1}^{\tau_{\max}}$ are
uniquely identified \citep{hyvarinen2010estimation}. 
The Gaussian-noise restriction in 
Proposition~\ref{prop:lingauss} is replaced by non-Gaussianity, 
which restores full DAG identifiability via independent component 
analysis on the residuals.
\end{proposition}

\paragraph{Nonlinear power-transform (recursive form).}
The library's nonlinear variant applies a sign-preserving saturating 
distortion $X_i \leftarrow \mathrm{sign}(Y_i)\,|Y_i|^{0.5}$ where 
$Y_i = \sum_j W_{ji} X_j + \sum_{\tau,j} (A_\tau)_{ji} X_{t-\tau,j} + N_{t,i}$. 
Because the transformed value is read by downstream variables as a 
parent in the same time slice, the resulting model is a recursive 
composition of distortions rather than a single-application 
post-nonlinear model. We do not establish identifiability for this 
recursive form; it is included to test method robustness to 
non-Gaussian, non-additive structure, not as an identifiability 
benchmark. We treat configurations using this mechanism as 
identifiability-unknown; the released core generators do not attach a 
machine-readable tag to them. Users requiring 
provably identifiable nonlinear DGPs should instead select the 
single-application post-nonlinear or additive-noise mechanisms, both 
of which are also supported.

\subsection{Counterfactual twin generation}
\label{sec:counterfactuals}

Traditional causal effect estimation benchmarks provide separate datasets 
for treatment and control groups, making individual treatment effect (ITE) 
evaluation impossible. \libname{} resolves this via cached-noise rollouts:

\begin{itemize}[leftmargin=*, itemsep=2pt, topsep=2pt, parsep=0pt]
    \item \textbf{Factual generation:} For each $(t, i)$, generate 
    $N_{t,i} \sim \mathcal{D}_i$, store it, and compute 
    $X_{t,i} = f_i(\text{pa}(X_{t,i}), N_{t,i})$.
    \item \textbf{Intervention:} Specify $\text{do}(X_{\tau,i} = c)$ at 
    intervention time $\tau$.
    \item \textbf{Counterfactual rollout:} Re-execute the forward pass from 
    $\tau$ onward, using the \emph{cached} noise $\bN$ but replacing 
    $X_{\tau,i}$ with $c$.
\end{itemize}

Because the stochastic component is held fixed, $\text{ITE}_{t,j}$ is 
exact. This enables evaluation metrics that require ground-truth individual 
effects: PEHE (Precision in Estimation of Heterogeneous Effects) and 
$\epsilon$-ATE (average treatment effect error).

\section{Experimental protocol}
\label{sec:experiments}

\subsection{Benchmark configuration}

We instantiate a tractable factorial design spanning fundamental difficulty 
axes:

{\setlength{\parskip}{0pt plus 1pt}
\paragraph{Graph configurations.}
Nine templates: $(d,\tau_{\max}) \in \{4,6,8\} \times \{2,3,4\}$. Each 
graph is hand-curated to include contemporaneous edges, lagged edges 
across several lags, and collider structures; all nine are acyclic. Sparsity is calibrated to maintain $\approx d$ edges per 
graph to avoid trivial or saturated regimes.

\paragraph{DGP variants.}
Figure~\ref{fig:sample-data} in Appendix~\ref{app:dgp-params} shows one graph rendered under four variants.
\begin{itemize}[leftmargin=*, itemsep=2pt, topsep=2pt, parsep=0pt]
    \item \textbf{Linear:} VAR with Gaussian noise. Identifies only 
    contemporaneous CPDAG.
    \item \textbf{Nonlinear:} Linear VAR followed by recursive sign-preserving 
    power transform $\mathrm{sign}(\cdot)|\cdot|^{0.5}$. Identifiability 
    not established for the recursive form.
    \item \textbf{Drift:} Linear VAR with linearly time-varying coefficients. 
    Tests robustness to non-stationarity.
    \item \textbf{MNAR:} Linear VAR with self-thresholded missingness 
    (top quantile of each variable masked). Tests methods on informative 
    missingness.
    \item \textbf{MCAR:} Linear VAR with uniform random missingness 
    (20\% per entry, independent of values). Serves as controlled 
    comparator to MNAR, isolating the cost of informativeness from the 
    cost of missingness itself.
    \item \textbf{Trend, Seasonal, Trend+Seasonal:} Linear VAR augmented 
    with additive exogenous deterministic components. The structural 
    graph is unchanged; these variants probe sensitivity to 
    non-stationary observational artifacts that do not enter the 
    structural equations.
    \item \textbf{Nonlinear+Trend+Seasonal:} Combination of the nonlinear 
    mechanism above with additive trend and seasonality.
\end{itemize}

\paragraph{Hyperparameter tuning.}
Every method with tunable hyperparameters undergoes Optuna~\citep{akiba2019optuna} TPE optimization with a budget of 30 trials 
per (method, cell) combination (VARLiNGAM and VARLiNGAM\_imp have no tunable parameters), optimized on oriented F1 against the 
ground-truth graph. Tuning is performed on a dedicated trajectory at 
seed $41$ that is held out of the reported evaluation set; no evaluation 
seed is ever seen by the tuning objective. Selected hyperparameters are 
cached at the (method, graph, variant) level and reused across all 
evaluation seeds for that cell, isolating tuning from the variance 
attributed to noise realizations. Two methods with prohibitive runtime, 
PCMCI-GPDC and DYNOTEARS, are tuned with a reduced budget of 5 trials. 
We note that the tuning objective is an oracle: it uses the ground-truth 
graph, which is unavailable in application. Reported numbers for tuned methods are therefore 
optimistic relative to what is attainable without the graph.

The benchmark fixes trajectory length $T{=}1000$, standard Gaussian 
innovations $\varepsilon_{t,i} \sim \mathcal{N}(0,1)$ shared across all 
variables, structural weights drawn from $\mathcal{U}(0.3, 0.8)$ with random 
sign and rejection-sampled to spectral radius $\rho < 0.95$, sparsity 
calibrated to $\approx d$ edges per graph, tuning seed $41$ and evaluation 
seeds $\{42, \ldots, 51\}$. Variant-specific parameters 
(drift slope, nonlinear transform, trend/seasonal activation, 
missingness fractions, CATE assignment policy) are tabulated in full 
in Appendix~\ref{app:dgp-params}.

\paragraph{Evaluation protocol.}
For each of the nine predefined graphs, the structural weights $W_0$ 
and $\{A_\tau\}$ are sampled once using a graph-derived weight seed 
and held fixed across all evaluation seeds, all DGP variants, and all 
methods. The ten evaluation seeds (42--51) vary only the noise 
realization $\{N_{t,i}\}$ within each fixed (graph, variant) cell. 
Reported standard deviations therefore reflect noise variance under 
held-fixed weights, not weight variance under held-fixed noise. This 
protocol is a deliberate design choice: it isolates method 
sensitivity to the stochastic component from confounding due to 
weight realization. A complementary weight-sensitivity analysis 
varying weights at fixed noise is supported by the framework but 
deferred to future work.

\paragraph{Baselines.}
We evaluate eight causal discovery methods spanning the four 
established paradigm classes (Granger-style, constraint-based, 
score-based, functional) and four CATE estimators covering 
representation-learning, structured-information, and meta-learning 
families. Discovery methods comprise VARLiNGAM~\citep{hyvarinen2010estimation} 
and its mean-imputation variant VARLiNGAM\_imp, PCMCI~\citep{runge2019detecting}
with both ParCorr and GPDC conditional independence tests, DYNOTEARS~\citep{pamfil2020dynotears}, 
CD-NOD~\citep{huang2020causal}, CUTS+~\citep{cheng2024cuts}, and a 
linear F-test Granger~\citep{granger1969investigating} baseline 
(GrangerF). CATE estimators comprise TARNet~\citep{shalit2017estimating}, 
FlexTENet~\citep{curth2021inductive}, SNet~\citep{curth2021nonparametric},
and X-Learner~\citep{kunzel2019metalearners}.

A unified wrapper interface (\texttt{fit(X, mask)} with declared 
\texttt{supports\_missing} flags) lets us treat heterogeneous 
methods uniformly: when a method declares it cannot accept missing 
inputs, the runner records a structured \texttt{skipped\_no\_missing\_support} 
status on MCAR and MNAR cells rather than silently emitting an empty 
graph. PCMCI-GPDC's GP-kernel cost restricts it to $d{=}4$ cells, 
with $d \in \{6, 8\}$ reported as \texttt{skipped\_compute\_budget}; 
CD-NOD targets contemporaneous causal structure under 
non-stationarity and does not detect lagged edges, which directly 
explains the low recall reported in Table~\ref{tab:main-discovery-obs}.
Per-method details appear in Appendix~\ref{app:baselines}.

\paragraph{Metrics.}
\textit{Discovery:} Structural Hamming Distance (SHD), normalized SHD 
(SHD / max possible edges), skeleton F1, orientation F1, true-positive 
rate (TPR), false-discovery rate (FDR), precision, recall, and AUPRC. 
\textit{CATE:} PEHE (Precision in Estimation of Heterogeneous Effects) 
and runtime.

}

\subsection{Causal discovery results}

Tables~\ref{tab:main-discovery-obs} and~\ref{tab:main-discovery-miss} present aggregate performance across observational and missingness DGP variants respectively. The ranking is strikingly DGP-dependent for some method classes and remarkably stable for others, a contrast we explore below.

\begin{table}[t]
\centering
\caption{Causal discovery performance on Linear and Nonlinear DGP 
variants. Mean ($\pm$ std) over 90 runs per cell (9 graphs $\times$ 
10 seeds), filtered to \texttt{status=ok} rows. Lower SHD and FDR 
are better; higher TPR is better. Bold marks the best value in 
each column. PCMCI-GPDC values reflect $d{=}4$ cells only ($n{=}30$ 
versus $n{=}90$ for other methods). Performance on Drift is 
within 0.08 SHD of Linear for every method; full 
results in Appendix~\ref{app:drift-table}.}
\label{tab:main-discovery-obs}
\footnotesize\setlength{\tabcolsep}{4.5pt}
\begin{tabular}{lcccccc}
\toprule
 & \multicolumn{3}{c}{Linear} & \multicolumn{3}{c}{Nonlinear} \\
 & SHD & TPR & FDR & SHD & TPR & FDR \\
\midrule
DYNOTEARS & \textbf{0.02 $\pm$ 0.05} & \textbf{1.00 $\pm$ 0.02} & \textbf{0.01 $\pm$ 0.04} & \textbf{0.02 $\pm$ 0.05} & \textbf{0.99 $\pm$ 0.03} & \textbf{0.01 $\pm$ 0.03} \\
PCMCI\_GPDC & 0.38 $\pm$ 0.17 & 0.67 $\pm$ 0.12 & 0.05 $\pm$ 0.11 & 0.41 $\pm$ 0.15 & 0.67 $\pm$ 0.12 & 0.07 $\pm$ 0.13 \\
VARLiNGAM & 0.52 $\pm$ 0.40 & 0.80 $\pm$ 0.17 & 0.27 $\pm$ 0.20 & 0.50 $\pm$ 0.41 & 0.79 $\pm$ 0.22 & 0.25 $\pm$ 0.20 \\
VARLiNGAM\_imp & 0.52 $\pm$ 0.40 & 0.80 $\pm$ 0.17 & 0.27 $\pm$ 0.20 & 0.50 $\pm$ 0.41 & 0.79 $\pm$ 0.22 & 0.25 $\pm$ 0.20 \\
PCMCI\_ParCorr & 0.45 $\pm$ 0.14 & 0.65 $\pm$ 0.07 & 0.12 $\pm$ 0.13 & 0.42 $\pm$ 0.12 & 0.65 $\pm$ 0.07 & 0.08 $\pm$ 0.10 \\
CUTS+ & 0.48 $\pm$ 0.24 & 0.87 $\pm$ 0.31 & 0.36 $\pm$ 0.24 & 0.44 $\pm$ 0.17 & 0.94 $\pm$ 0.12 & 0.27 $\pm$ 0.09 \\
GrangerF & 1.83 $\pm$ 0.58 & 0.65 $\pm$ 0.07 & 0.67 $\pm$ 0.09 & 1.83 $\pm$ 0.59 & 0.65 $\pm$ 0.07 & 0.67 $\pm$ 0.10 \\
CDNOD & 1.03 $\pm$ 0.10 & 0.03 $\pm$ 0.07 & 0.92 $\pm$ 0.17 & 1.02 $\pm$ 0.07 & 0.02 $\pm$ 0.05 & 0.94 $\pm$ 0.16 \\
\bottomrule
\end{tabular}

\end{table}

\begin{table}[t]
\centering
\caption{Causal discovery performance on missingness DGP variants. 
Only the two methods declaring \texttt{supports\_missing=True} are 
shown; the remaining six methods were skipped on these variants 
(see Appendix~\ref{app:failures}). Bold marks the best value in 
each column.}
\label{tab:main-discovery-miss}
\footnotesize\setlength{\tabcolsep}{4.5pt}
\begin{tabular}{lcccccc}
\toprule
 & \multicolumn{3}{c}{MNAR} & \multicolumn{3}{c}{MCAR} \\
 & SHD & TPR & FDR & SHD & TPR & FDR \\
\midrule
VARLiNGAM\_imp & 0.96 $\pm$ 0.37 & 0.52 $\pm$ 0.26 & 0.47 $\pm$ 0.23 & 0.79 $\pm$ 0.43 & 0.64 $\pm$ 0.26 & 0.40 $\pm$ 0.24 \\
CUTS+ & \textbf{0.94 $\pm$ 0.45} & \textbf{0.73 $\pm$ 0.18} & \textbf{0.42 $\pm$ 0.16} & \textbf{0.67 $\pm$ 0.43} & \textbf{0.94 $\pm$ 0.10} & \textbf{0.35 $\pm$ 0.13} \\
\bottomrule
\end{tabular}

\end{table}

{\setlength{\parskip}{0pt plus 1pt}

\noindent\textbf{Key findings.}

\textbf{DYNOTEARS dominates clean conditions} with F1 $\approx 0.99$ on 
Linear and Nonlinear and 0.98 on Drift 
(Appendix~\ref{app:drift-table}; SHD 0.02, 0.02 and 0.04). Recovery on the 
Nonlinear variant is driven by the recursive sign-power transform's 
strict monotonicity and bounded amplitude rather than general nonlinear 
robustness, as we examine in Section~\ref{sec:discussion} and 
Appendix~\ref{app:nonlin-monotonicity}. DYNOTEARS does not support 
missing inputs and is therefore absent from MCAR and MNAR.

\textbf{CUTS+ is the strongest method on missing data} (F1 = 0.75 on 
MCAR, 0.62 on MNAR), completing all 90 cells under each and beating 
VARLiNGAM\_imp on SHD, TPR, and FDR under both. The 0.14-point 
gap quantifies the cost of informativeness for a method that handles 
missingness structurally rather than through input rejection.

\textbf{Linear conditional-independence methods are insensitive to 
monotonic and additive perturbations.} At evaluation seed 42, PCMCI-ParCorr and GrangerF return the same SHD 
and predicted-edge count on the Linear and Nonlinear versions of a graph 
in 15 of 18 (graph, method) pairs.

\textbf{VARLiNGAM\_imp completes every cell} (F1 = 0.61 on MCAR, 0.50 on 
MNAR) after a wrapper fix; the 171 failures of our first run biased its 
earlier scores upward (Appendix~\ref{app:failures}).

\textbf{CD-NOD's near-zero recall} (TPR 0.02 to 0.03) reflects CD-NOD's design as a contemporaneous-structure recovery method, not a method failure. 
\textbf{GrangerF serves as the adversarial floor} (F1 $\approx 0.43$, 
FDR $\approx 0.67$).

Per-dimension performance breakdowns appear in Appendix~\ref{app:dim-scaling}: 
DYNOTEARS scales favorably (SHD essentially flat across $d \in \{4, 6, 8\}$), 
while PCMCI-ParCorr and GrangerF degrade at $d = 8$; PCMCI-GPDC's 
GP-kernel cost places $d \in \{6, 8\}$ outside our compute budget.

\paragraph{Failure modes and method-level instability.}
Of 6{,}480 attempted (method, graph, variant, seed) combinations, 4{,}809
succeeded; 960 were skipped because the method declared
\texttt{supports\_missing=False} on a missingness variant; 540 were skipped
on PCMCI-GPDC at $d > 4$ by the compute-budget guard; and none raised an
exception. Our first run had 171 exceptions (2.6\%), all in
VARLiNGAM\_imp, from a wrapper defect; this paper reports a full regeneration with the 
wrapper fixed (Appendix~\ref{app:failures}).

}

\subsection{Varsortability and the marginal-variance shortcut}
\label{sec:varsortability}

Additive-noise generators with homoscedastic innovations and coefficients
bounded away from zero accumulate variance along structural paths.
\citet{reisach2021beware} show that the resulting \emph{varsortability} $v$,
the fraction of directed paths running from lower to higher marginal variance,
can make a benchmark solvable by sorting variables by variance and regressing,
without causal reasoning. We measure $v$ rather than assume it away.

Table~\ref{tab:varsort} and Figure~\ref{fig:varsortability} in
Appendix~\ref{app:dgp-params} report the audit per graph; the headline
numbers follow. Contemporaneous varsortability on the raw benchmark averages $0.93$
($0.5$ is uninformative, $1.0$ perfectly sortable), so we add two controls.
\textsc{SortnRegress} orders contemporaneous variables by increasing marginal
variance and fits an adaptive lasso; \textsc{RandomRegress} is the identical
estimator with a random order; lagged variables are candidate parents for
both, since time resolves their direction. On raw data \textsc{SortnRegress}
reaches $F_1 = 0.93$ against DYNOTEARS at $0.99$, so most of the linear
variant's apparent difficulty is carried by marginal variance. Standardizing
drives $v$ to exactly $0.5$ while leaving the graph unchanged; DYNOTEARS falls
to $0.82$, \textsc{SortnRegress} converges to \textsc{RandomRegress} ($0.81$
against $0.79$, the latter invariant across conditions), the margin of
DYNOTEARS over the best competing method shrinks from $0.23$ to $0.05$, and
re-tuning separately for the standardized condition on the held-out trajectory
recovers less than $0.01$. Across the nine DGP variants the effect is material
($>0.05$) on six, largest on seasonal ($0.79$ to $0.46$), absent on the trend
variants whose exogenous variance dilutes $v$, and exactly zero for every
scale-invariant method (PCMCI-ParCorr, VARLiNGAM, GrangerF), so the artifact
is method-selective. Figure~\ref{fig:recovered} in
Appendix~\ref{app:dgp-params} shows the effect on one cell, $F_1$ $1.00$ raw
and $0.44$ standardized.

Two conclusions follow. First, the near-perfect linear-variant scores in
Table~\ref{tab:main-discovery-obs} overstate what these methods recover from
structure alone, with the score-based method affected most, consistent with
\citet{reisach2021beware}. Second, the effect is bounded rather than total:
the contemporaneous sub-graphs are shallow, so one to three of the four to
eight edges per graph are exposed to the shortcut, and standardized
performance remains well above chance. \libname{} now ships standardization
as a first-class DGP axis (the \texttt{\_std} variants), records $v$ in the
metadata of every generated cell, and includes both controls in the default
baseline set, so any result can be read against the shortcut it might be
exploiting. Appendix~\ref{app:timegraph} applies the same instrument to an
independently authored equation family and finds the reverse regime, high
difficulty with no shortcut.

\subsection{Counterfactual estimation results}

Table~\ref{tab:cate-main} presents CATE estimator performance. The twin 
generation protocol enables PEHE evaluation against ground-truth 
individual treatment effects, a measurement unavailable in conventional 
benchmarks that compare predictions to conditional expectations.

\begin{table}[t]
\centering
\caption{CATE estimator performance aggregated by DGP variant. Mean ($\pm$ std) over 90 runs per cell (9 graphs $\times$ 10 seeds). PEHE: Precision in Estimation of Heterogeneous Effects; lower is 
better. Runtime is wall-clock per fit, measured under four parallel processes (Appendix~\ref{app:compute}); compare ratios only.}
\label{tab:cate-main}
\small
\begin{tabular}{lcccc}
\toprule
 & \multicolumn{2}{c}{CATE Linear} & \multicolumn{2}{c}{CATE Nonlinear} \\
 & PEHE & Runtime (s) & PEHE & Runtime (s) \\
\midrule
TARNet & \textbf{0.491 $\pm$ 0.173} & 1.639 $\pm$ 0.349 & \textbf{0.427 $\pm$ 0.135} & 1.639 $\pm$ 0.351 \\
SNet & 0.674 $\pm$ 0.220 & 11.571 $\pm$ 1.436 & 0.509 $\pm$ 0.129 & 12.444 $\pm$ 1.988 \\
FlexTENet & 0.663 $\pm$ 0.201 & 6.385 $\pm$ 1.244 & 0.518 $\pm$ 0.129 & 6.636 $\pm$ 1.302 \\
X-Learner & 0.757 $\pm$ 0.155 & 1.380 $\pm$ 0.078 & 0.617 $\pm$ 0.129 & 1.374 $\pm$ 0.078 \\
\bottomrule
\end{tabular}

\end{table}

\noindent\textbf{Key findings.} 
\textbf{TARNet wins on both variants} (PEHE = 0.49 Linear, 0.43 Nonlinear) with 
a runtime of about 1.6 seconds, comparable only to X-Learner; the representation-balancing objective 
fits the heterogeneous effect structure \libname{} produces under randomized assignment. 
\textbf{All methods improve from Linear to Nonlinear} (PEHE drops 13--25\% across 
all four estimators). The pattern reflects the saturating sign-power transform 
compressing the range of treatment-effect modifiers, narrowing effective 
heterogeneity, a mechanism-level interaction that aggregate single-DGP benchmarks 
would not surface. \textbf{Runtime-accuracy tradeoffs are stark}: TARNet and 
X-Learner complete in under 2 seconds; FlexTENet takes about 6.5 seconds and trails TARNet by 0.09 to 0.17 PEHE; SNet's 12 seconds buys no PEHE 
advantage over FlexTENet, which may reflect a mismatch between its 
orthogonal-information design and \libname{}'s relatively short 
($T = 1000$) sequences. \textbf{Method ranking is preserved across variants} 
(TARNet $<$ SNet $\approx$ FlexTENet $<$ X-Learner on both); deeper architectures 
provide little additional benefit once the average effect is captured.

Per-dimension PEHE patterns on the Linear DGP (Appendix~\ref{app:dim-scaling}) are mixed: 
X-Learner's PEHE falls with $d$, TARNet's does not rise monotonically, and SNet's and 
FlexTENet's rise.

\section{Discussion}
\label{sec:discussion}

\noindent\textbf{Inspecting individual failures reveals patterns hidden 
by aggregate metrics.} CD-NOD's TPR of 0.02 to 0.03 reflects the algorithm's design: CD-NOD recovers contemporaneous causal structure under non-stationarity and does not detect lagged edges, so when evaluated against ground-truth graphs that contain both, its measured recall undercounts what the method is designed to find. This is a benchmark-versus-method scope mismatch rather than a method failure, and we report it transparently. GrangerF's FDR $\approx 0.67$ is consistent with Granger causality's known inability to distinguish direct from mediated effects. PCMCI-ParCorr and GrangerF return unchanged SHD and edge counts between Linear and Nonlinear in most graphs (Section~\ref{sec:experiments}), consistent with linear conditional-independence tests being insensitive to monotonic 
distortion.

\noindent\textbf{DYNOTEARS' nonlinear-DGP performance reflects 
monotonicity plus signal-amplitude stability.} Although DYNOTEARS 
optimizes a linear SEM, it achieves near-perfect F1 on the Nonlinear 
variant. The recursive sign-power transform $\text{sign}(x)|x|^{0.5}$ 
is strictly monotonic and bounded by the square-root saturation, so 
the sign of every partial derivative is preserved and signal 
magnitudes remain well-conditioned for linear regression. We probe 
this boundary empirically in Appendix~\ref{app:nonlin-monotonicity}: 
a non-monotonic alternative ($\sin x$) drops F1 from 0.99 to 0.88
and raises mean SHD from 0.11 to 1.14. The Nonlinear variant remains 
identifiability-unknown (Section~\ref{sec:identifiability}); we report this as bounded robustness 
evidence under monotonic distortion rather than identifiability 
evidence.

\noindent\textbf{Missing values defeat methods regardless of 
mechanism.} Six of eight discovery methods cannot accept NaN inputs 
at all. Of the two that opt in, CUTS+ completes all 90 cells under 
both MCAR and MNAR (F1 0.75 and 0.62). We attribute the 0.14-point gap, without having tested it, to 
neural imputation extrapolating poorly into MNAR's systematically 
masked upper quantiles. VARLiNGAM\_imp also completes every cell (F1 0.61 and 0.50) and trails CUTS+ 
on both; its earlier failures were a wrapper defect, not missingness 
(Appendix~\ref{app:failures}). What the benchmark does show is that input-validation barriers, 
not identifiability, keep most methods from running on missing data at 
all; \libname{}'s MCAR variant exists to make that distinction observable.

\noindent\textbf{Counterfactual evaluation reveals architectural mismatches.} Method ranking is ordinally stable across DGP variants (TARNet $<$ SNet $\approx$ FlexTENet $<$ X-Learner on both Linear and Nonlinear), in contrast to the discovery side where DGP-dependence reshuffled rankings. Architectural complexity does not translate to PEHE gains: SNet's orthogonal-information design takes about seven times TARNet's runtime to produce worse PEHE on both variants, a mismatch flat ATE-error reporting would hide.

\noindent\textbf{Dimensionality is not uniformly detrimental.} 
On the Linear DGP, X-Learner's PEHE falls with $d$ (0.84, 0.82, 0.61 for $d = 4, 6, 8$) 
TARNet's does not rise monotonically (0.44, 0.59, 0.44), and SNet's and FlexTENet's 
rise (Appendix~\ref{app:dim-scaling}). We have not tested why, and with three dimensions and 30 runs 
per cell we treat it as a pattern to examine. This underscores why compositional frameworks that 
isolate the causal effect of each axis are valuable.
\section{Limitations and broader impact}

\subsection{Limitations}

\noindent\textbf{What synthetic data can and cannot establish.} A synthetic
generator fixes the mechanism class by construction, and that class tends to
resemble the model forms prevailing when the generator is written, so results
on \libname{} are evidence about behavior under a stated DGP, not about
empirical data; Appendix~\ref{app:timegraph} quantifies this family
dependence on an independently authored DGP. Synthetic evaluation is
nonetheless required rather than optional. Structure recovery is scored
against the true graph, which no observational corpus supplies and expert
annotation only approximates, and the counterfactual outcomes needed to score
treatment-effect estimation exactly are never observed. The right reading is
therefore diagnostic rather than confirmatory, one leg of an evaluation that
also includes real data, and we provide export adapters to
\texttt{tigramite} and \texttt{causal-learn} for that purpose.
Section~\ref{sec:varsortability} is a worked instance of the risk, a
generator artifact that no aggregate metric surfaces, found only by measuring
for it directly; we expect others remain, and we ship the controls partly so
the next one is easier to catch.

\noindent\textbf{Scope.} The reference benchmark spans $d \in \{4, 6, 8\}$,
with PCMCI-GPDC restricted to $d{=}4$ on compute grounds; larger graphs
(Appendix~\ref{app:scalability}) and the spatio-temporal module are supported
but outside the reference benchmark. Deferred baselines and implementation notes
are in Appendix~\ref{app:deferred}.

\noindent\textbf{Identifiability.} Configurations whose identifiability 
is not established (recursive nonlinear power-transform; non-monotonic 
post-nonlinear distortions; simultaneous MNAR on confounded variables; 
feedback without lag separation) are treated as 
identifiability-unknown. Only the spatio-temporal module currently emits 
identifiability tags and construction-time warnings; adding them to the 
core generators is planned. Performance on these configurations is 
robustness evidence, not identifiability evidence.

\noindent\textbf{Protocol.} Standard deviations capture noise-realization 
variance under weights held fixed per graph; weight-realization 
sensitivity is supported but deferred. Hyperparameter selection uses an 
oracle objective (F1 against the ground-truth graph) on a held-out 
trajectory, so reported figures are optimistic relative to what is 
attainable without the graph. CATE assignment is 
Bernoulli$(0.5)$, isolating function approximation from deconfounding; 
confounded-assignment variants are scheduled for the next release. 
The \texttt{CounterfactualSuite} presumes causal sufficiency of the 
simulator's variable set, which holds by construction but does not 
transfer to empirical counterfactual evaluation. The reference results are deterministic given their seeds on one machine, which we checked by rerunning selected cells in separate processes (Appendix~\ref{app:failures}); we have not tested other machines.

\subsection{Broader impact}
\libname{} is a research benchmark; it does not deploy decisions, control
physical systems, or process personal data, and we see no misuse pathway
specific to it. Explicit identifiability documentation should reduce methods
being credited for performance on unidentifiable configurations, and
protocol-level standardization (uniform tuning, structured failure reporting,
controlled DGP variation) improves rigor in follow-up work. A Datasheet
(Appendix~\ref{app:datasheet}) and Croissant metadata accompany the artifact,
under Apache-2.0 (library) and CC-BY-4.0 (datasets).



\bibliographystyle{unsrtnat}
\clearpage
\section*{Acknowledgments and Disclosure of Funding}
We thank the NeurIPS reviewers and the area chair, whose questions during the
discussion period led directly to the varsortability audit of
Section~\ref{sec:varsortability}, the scalability study of
Appendix~\ref{app:scalability}, and the corrected tuning protocol.
Experiment execution and manuscript editing were assisted by LLM-based
tools, as disclosed in the submission form; the authors are responsible for
all content.

Research reported in this publication was supported by the NSF iHARP: HDR Institute
for Harnessing Data and Model Revolution in the Polar Regions (Award \#2118285) and by the
National Institute on Minority Health and Health Disparities of the National Institutes of
Health (Award \#R01MD019814). The views expressed in this work are solely the responsibility
of the authors and do not necessarily represent the official views of the NSF or the NIH.
Endorsement by the Federal Government should not be inferred.

The authors declare no competing financial interests related to this work.

\bibliography{references}

\appendix
\let\appendixsection\section
\renewcommand{\section}{\FloatBarrier\appendixsection}

\clearpage
\section{Reference benchmark: DGP parameters and illustrations}
\label{app:dgp-params}

\begin{table}[!htbp]
\centering
\caption{DGP parameters fixed across the reference benchmark (referenced from Section~\ref{sec:experiments}). The "Variant" column lists where each parameter applies; "all" indicates the parameter is shared by every DGP variant. "T+S" abbreviates "Trend+Seasonal"; "N+T+S" abbreviates "Nonlinear+Trend+Seasonal".}
\label{tab:dgp-params}
\small
\resizebox{\textwidth}{!}{%
\begin{tabular}{l l l}
\toprule
\textbf{Parameter} & \textbf{Value} & \textbf{Variant} \\
\midrule
Trajectory length $T$ & 1000 & all \\
Within-graph variables $d$ & $\{4, 6, 8\}$ & all \\
Maximum lag $\tau_{\max}$ & $\{2, 3, 4\}$ & all \\
Edges per graph & $\approx d$ (sparsity calibrated) & all \\
Innovation distribution & Gaussian, $\mathcal{N}(0,1)$, homoscedastic & all \\
Structural weight magnitude & $\sim \mathcal{U}(0.3, 0.8) \cdot \pm 1$ & all \\
Stability constraint & companion spectral radius $\rho < 0.95$ & all \\
Weight seed & \texttt{md5(graph\_id)}, fixed per graph & all \\
Drift coefficient slope $\beta_{\text{drift}}$ & $0.5$ (linear in $t/T$) & Drift \\
Nonlinear transform & $\text{sign}(\cdot)|\cdot|^{0.5}$ (recursive) & Nonlinear, N+T+S, CATE Nonlinear \\
Trend activation per variable & $p{=}0.5$ (Bernoulli) & Trend, T+S, N+T+S \\
Trend amplitude $\alpha_i$ (when active) & $\sim \mathcal{U}(0.5, 2.0) \cdot \pm 1$ & Trend, T+S, N+T+S \\
Active seasonal components per variable & $\{0, 1, 2\}$ with $p{=}(0.3, 0.4, 0.3)$ & Seasonal, T+S, N+T+S \\
Seasonal period candidates $P_k$ & $\{12, 24, 50, 168\}$ (feasible if $T \geq 3 P_k$) & Seasonal, T+S, N+T+S \\
Seasonal amplitudes $\beta_{i,k}$ & $\sim \mathcal{U}(0.3, 1.0)$ & Seasonal, T+S, N+T+S \\
Seasonal phases $\phi_{i,k}$ & random per component & Seasonal, T+S, N+T+S \\
MCAR missing fraction & 0.20 (per entry) & MCAR \\
MNAR threshold quantile & top 0.20 (per variable) & MNAR \\
CATE treatment type & binary, replaces treatment-node value at each $t$ & CATE Linear, CATE Nonlinear \\
CATE observational treatment & $\text{Bernoulli}(0.5)$ at each $t$ & CATE Linear, CATE Nonlinear \\
CATE counterfactual rollouts & $\mathbf{1}_T$ (always-on) and $\mathbf{0}_T$ (always-off) & CATE Linear, CATE Nonlinear \\
Tuning seed (held out) & $41$ & all \\
Evaluation seeds & $\{42, 43, \ldots, 51\}$ & all \\
\bottomrule
\end{tabular}}
\end{table}

\begin{figure}[!htbp]
\centering
\includegraphics[width=\textwidth]{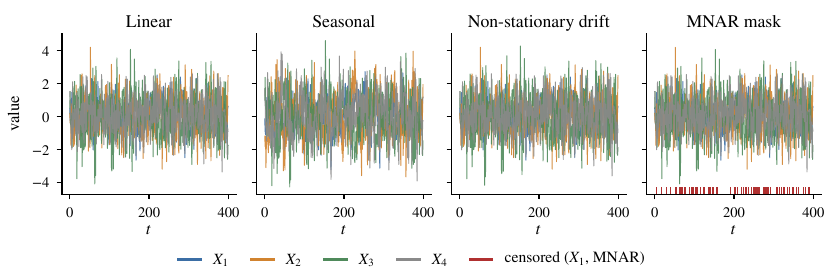}
\caption{Generated trajectories for graph \texttt{d4\_tau2} at seed 42 under
four DGP variants ($T{=}400$ shown for legibility). Graph, weights and noise realization are identical across
panels; only the varied axis differs. Red ticks in the rightmost panel mark
entries censored by the MNAR mask on $X_1$.}
\label{fig:sample-data}
\end{figure}

\begin{figure}[!htbp]
\centering
\includegraphics[width=\textwidth]{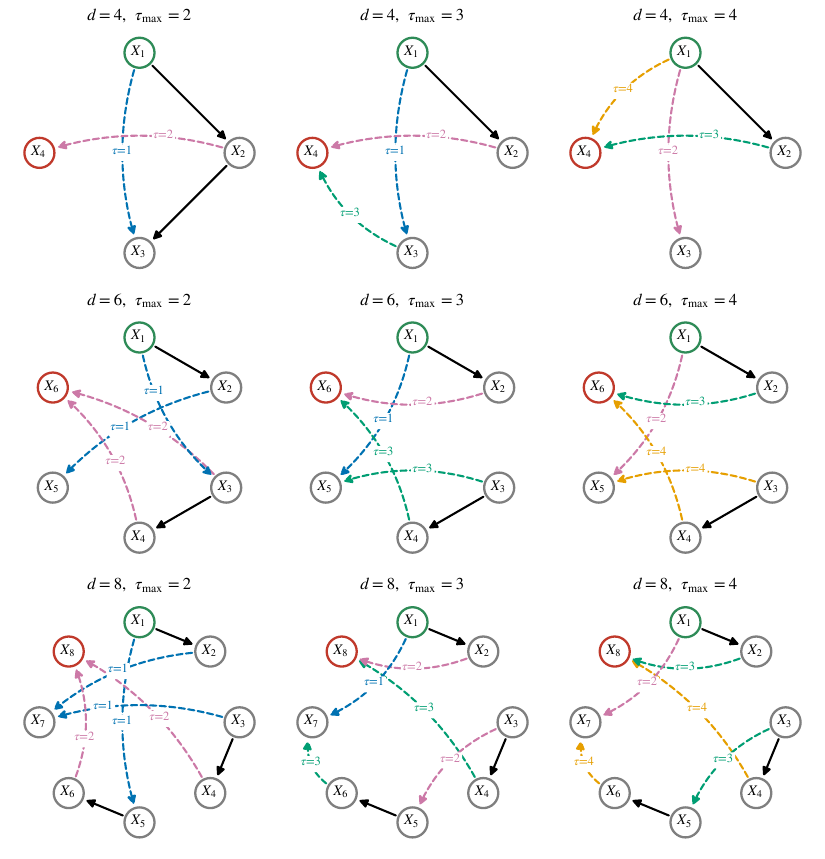}
\caption{Ground-truth window graphs for the nine reference templates (rows
$d$, columns $\tau_{\max}$). Solid black arrows are contemporaneous edges;
dashed arrows are lagged edges, colored and labeled by their lag $\tau$.
Green and red outlines mark the treatment and outcome nodes used by the CATE
variants. Sparsity is held at $\approx d$ edges per graph.}
\label{fig:ground-truth}
\end{figure}

\begin{figure}[!htbp]
\centering
\includegraphics[width=\textwidth]{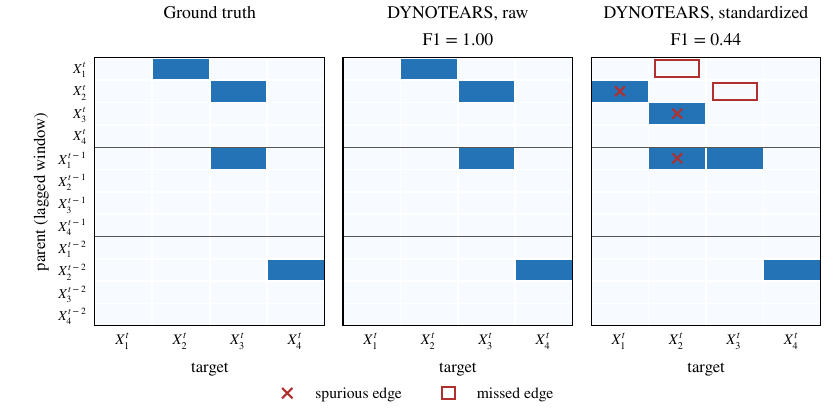}
\caption{Recovered window graph beside ground truth for \texttt{d4\_tau2},
linear variant, seed 42, identical hyperparameters. Rows are candidate parents
across the lag window, columns are targets at time $t$; crosses mark
spurious edges and open boxes mark missed edges. DYNOTEARS recovers the
graph exactly on raw data and misplaces a majority of edges once the
marginal-variance ordering is removed.}
\label{fig:recovered}
\end{figure}

\begin{figure}[!htbp]
\centering
\includegraphics[width=\textwidth]{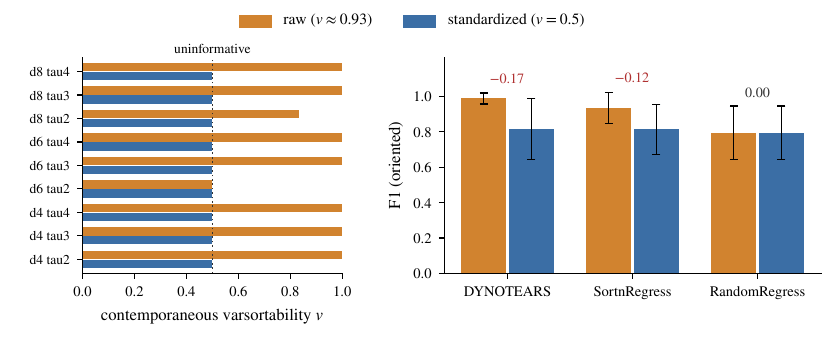}
\caption{Left: contemporaneous varsortability per graph, raw versus
standardized. Right: oriented F1 (nine graphs, ten seeds, fixed
hyperparameters and DGP); annotations give the F1 lost to standardization.}
\label{fig:varsortability}
\end{figure}

\begin{table}[!htbp]
\centering
\caption{Varsortability audit on the linear reference benchmark; oriented F1
over nine graphs $\times$ ten seeds with hyperparameters, weights and noise
realizations fixed across conditions. \textsc{SortnRegress} and
\textsc{RandomRegress} are controls, not discovery methods. The DYNOTEARS
settings come from an earlier tuning run of the same protocol and were not
recomputed after the final reference run.}
\label{tab:varsort}
\small
\begin{tabular}{lcccc}
\toprule
Condition & $v$ & DYNOTEARS & \textsc{SortnRegress} & \textsc{RandomRegress} \\
\midrule
Raw (as shipped) & 0.93 & \textbf{0.99} & 0.93 & 0.79 \\
Standardized     & 0.50 & 0.82 & 0.81 & 0.79 \\
\midrule
$\Delta$         & $-0.43$ & $-0.17$ & $-0.12$ & $0.00$ \\
\bottomrule
\end{tabular}

\end{table}

\section{Baseline methods}
\label{app:baselines}

We provide brief descriptions of each baseline included in the 
reference benchmark, with implementation source, supported input 
modalities, and key assumptions. Hyperparameter search spaces appear 
in Appendix~\ref{app:hyperparameters}.

\subsection{Causal discovery baselines}

\noindent\textbf{VARLiNGAM} \citep{hyvarinen2010estimation}. 
Linear non-Gaussian SVAR estimator combining vector autoregression 
with Independent Component Analysis on the residuals. Identifies the 
full DAG (both contemporaneous and lagged) under the assumptions of 
linearity, causal sufficiency, acyclicity, and at most one Gaussian 
innovation per variable. Implementation: \texttt{lingam} package. 
Does not support missing inputs.

\noindent\textbf{VARLiNGAM\_imp.} VARLiNGAM with mean imputation 
applied as a preprocessing step before fitting. Included to enable 
participation in the MCAR and MNAR variants. The original run's failures 
for this wrapper are documented, with their cause, in 
Appendix~\ref{app:failures}.

\noindent\textbf{PCMCI-ParCorr} \citep{runge2019detecting}. 
Constraint-based two-stage algorithm using partial correlation as 
the conditional independence test. Recovers lagged causal links 
under causal sufficiency, faithfulness, and stationarity; the 
ParCorr test additionally assumes linear Gaussian dependencies. 
Implementation: \texttt{tigramite} package. Does not support 
missing inputs.

\noindent\textbf{PCMCI-GPDC} \citep{runge2019detecting}.
PCMCI with the GPDC test, which removes the conditioning set by Gaussian
process regression and tests the residuals with distance correlation,
accommodating nonlinear dependencies. 
Quadratic-in-$T$ kernel cost restricts applicability to $d{=}4$ in 
our compute budget. Implementation: \texttt{tigramite}. Does not 
support missing inputs.

\noindent\textbf{DYNOTEARS} \citep{pamfil2020dynotears}. 
Score-based continuous-optimization framework that jointly learns 
contemporaneous and lagged adjacency matrices via a smooth 
acyclicity constraint. Assumes linear SEMs; assumes neither 
Gaussian nor non-Gaussian noise specifically. Implementation:
vendored from the \texttt{causalnex} package. Does not support missing inputs.

\noindent\textbf{CD-NOD} \citep{huang2020causal}. 
Constraint-based discovery for 
heterogeneous and non-stationary data, treating the time index as a 
surrogate variable to recover contemporaneous causal structure. The 
algorithm is designed for contemporaneous edge recovery and does not 
detect lagged edges in its standard form. Implementation: 
\texttt{causal-learn}. Since our reference benchmark graphs contain 
both contemporaneous and lagged edges, CD-NOD's measured TPR 
$\approx$ 0.03 (Table~\ref{tab:main-discovery-obs}) reflects this scope mismatch rather than 
method failure.

\noindent\textbf{CUTS+} \citep{cheng2024cuts}. 
Joint neural imputation and causal discovery for irregularly sampled 
or missing-value time series. Uses message passing on a Granger 
causality graph with delayed supervision. Natively supports missing 
inputs (\texttt{supports\_missing=True}); one of the two methods 
participating in MCAR and MNAR cells in our benchmark.

\noindent\textbf{GrangerF.} Linear F-test baseline assessing 
Granger causality between every pair of variables at each lag, with 
a configurable significance threshold. Included as the adversarial 
floor: linear F-tests on marginal lagged dependencies cannot 
distinguish direct from mediated effects, and the resulting 
FDR $\approx 0.67$ confirms this expected behavior.

\subsection{CATE estimation baselines}

\noindent\textbf{TARNet} \citep{shalit2017estimating}. 
Two-headed neural network with a shared representation trunk and
separate treatment-specific heads. It is the variant of the
counterfactual regression framework without the integral probability
metric balancing penalty.

\noindent\textbf{SNet} \citep{curth2021nonparametric}. 
Multi-headed architecture with orthogonal information layers 
designed to disentangle prognostic, propensity-relevant, and 
treatment-effect-relevant signal. Higher parameter count than 
TARNet; designed to outperform under high-dimensional confounding.

\noindent\textbf{FlexTENet} \citep{curth2021inductive}. 
Flexible-architecture variant of two-headed CATE estimation with 
soft information-sharing between control and treatment heads via 
adaptive layer-sharing. Intermediate complexity between TARNet 
and SNet.

\noindent\textbf{X-Learner} \citep{kunzel2019metalearners}. 
Meta-learner that fits an outcome model for each treatment arm,
regresses the imputed individual treatment effects on covariates within
each arm, and combines the two estimates with propensity-score weights.
Model-agnostic; we instantiate 
the base learners as random forests (via EconML).

\subsection{Wrapper interface}

All causal discovery wrappers expose the unified signature 
\texttt{fit(X, mask=None) $\rightarrow$ AdjacencyResult}, with a 
declared boolean class attribute \texttt{supports\_missing}. CATE 
wrappers expose \texttt{fit(X, A, mask=None) $\rightarrow$ tau\_hat}. 
Methods declaring \texttt{supports\_missing=False} are skipped on 
MCAR and MNAR variants and recorded with status 
\texttt{skipped\_no\_missing\_support}, rather than silently 
returning empty graphs.

\subsection{Deferred baselines and reference implementation}
\label{app:deferred}
\textbf{Baseline coverage.} The benchmark covers eight discovery methods and
four CATE estimators across the four discovery paradigms. Recent methods
(NTS-NOTEARS \citep{sun2023ntsnotears}, stability-oriented temporal DAG
learners, CRN \citep{bica2020crn}, G-Net \citep{li2021gnet}, Causal
Transformer \citep{melnychuk2022causal}) require non-trivial wrapper adaptation
and are deferred to follow-up; the unified \texttt{fit(X, mask)} interface
lowers the cost of community-contributed baselines.

\textbf{Reference implementation.} The benchmark was produced by a monolithic
\texttt{generate\_datasets.py} for runtime efficiency; the user-facing OO API
in \texttt{dtcd.generators} provides the same capabilities. End-to-end
equivalence is verified per-cell rather than guaranteed by construction; a
unified single-path implementation is on the roadmap.

\section{Evaluation metrics}
\label{app:metrics}

We report all metrics over filtered \texttt{status=ok} cells 
unless otherwise noted; skipped and errored cells are tabulated 
separately in Appendix~\ref{app:failures}.

\subsection{Causal discovery metrics}

\noindent\textbf{Structural Hamming Distance (SHD).} 
The number of edge additions, deletions, and reversals required to 
transform the predicted adjacency $\hat{A}$ into the ground-truth 
adjacency $A^*$:
\begin{equation}
\text{SHD}(\hat{A}, A^*) = \sum_{i,j,\tau} \mathbb{1}\!\left[\hat{A}_{i,j,\tau} \neq A^*_{i,j,\tau}\right] - \tfrac{1}{2}\,r(\hat{A}, A^*),
\end{equation}
where $r(\cdot, \cdot)$ counts reversed edges (each contributes one 
to the raw symmetric difference but is conventionally weighted as a 
single edit). Lower is better.

\noindent\textbf{Normalized SHD.} 
$\text{SHD}(\hat{A}, A^*) / E_{\max}$, where 
$E_{\max} = d^2 (\tau_{\max} + 1)$ is the maximum possible number 
of directed edges in the unrolled time-graph representation. 
Reported alongside raw SHD to enable cross-graph comparison at 
different $d$ and $\tau_{\max}$.

\noindent\textbf{Skeleton F1 / Orientation F1.} 
Skeleton F1 is the F1 score on the undirected adjacency 
($\hat{A}_{i,j,\tau} \vee \hat{A}_{j,i,\tau}$ vs.\ ground truth), 
ignoring edge direction. Orientation F1 is the F1 score on directed 
edges, requiring both presence and direction. Higher is better. 
Tables reporting "F1" without qualification refer to skeleton F1.

\noindent\textbf{True Positive Rate (TPR) and False Discovery Rate (FDR).}
\begin{equation}
\text{TPR} = \frac{|\hat{E} \cap E^*|}{|E^*|}, \qquad
\text{FDR} = \frac{|\hat{E} \setminus E^*|}{|\hat{E}|},
\end{equation}
where $\hat{E}$ and $E^*$ are the predicted and ground-truth edge 
sets respectively. TPR is recall; FDR is the complement of precision. 
Higher TPR and lower FDR are better.

\noindent\textbf{AUPRC.} 
Area under the precision-recall curve, computed by sweeping the 
edge-confidence threshold across the predicted continuous adjacency 
matrix prior to thresholding. Captures method behavior across the 
full operating range, complementing the threshold-dependent F1 / TPR 
/ FDR metrics.

\subsection{Treatment-effect estimation metrics}

\noindent\textbf{PEHE (Precision in Estimation of Heterogeneous Effects).} 
Root mean squared error between estimated and true individual 
treatment effects, averaged across timesteps and trajectories:
\begin{equation}
\text{PEHE} = \sqrt{\frac{1}{T \cdot N} \sum_{t, n} \left( \hat{\tau}_{t,n} - \tau^*_{t,n} \right)^2},
\end{equation}
where $\tau^*_{t,n} = X^{\text{do}(A=1)}_{t,n} - X^{\text{do}(A=0)}_{t,n}$ 
is the ground-truth ITE recovered exactly from cached-noise 
counterfactual rollouts, and 
$\hat{\tau}_{t,n}$ is the estimator's prediction. Lower is better. 
PEHE evaluation against ground-truth ITEs is the metric \libname{}'s 
counterfactual-twin construction was specifically designed to 
enable.

\noindent\textbf{$\epsilon$-ATE.} 
Absolute error in the estimated average treatment effect:
\begin{equation}
\epsilon\text{-ATE} = \left| \frac{1}{T \cdot N} \sum_{t, n} \hat{\tau}_{t,n} - \frac{1}{T \cdot N} \sum_{t, n} \tau^*_{t,n} \right|.
\end{equation}
Reported in supplementary tables; PEHE is preferred in main results 
because it captures heterogeneity rather than only the average effect.

\noindent\textbf{Runtime.} 
Wall-clock time per \texttt{fit} call, in seconds, on the reference 
hardware described in Appendix~\ref{app:compute}. Reported alongside 
PEHE to surface accuracy-cost tradeoffs.

\section{Hyperparameter search spaces}
\label{app:hyperparameters}

For each baseline we report the Optuna search space actually used by the released code. Each study runs 30 TPE trials per (method, graph, variant) cell on the held-out trajectory (seed 41), except PCMCI-GPDC and DYNOTEARS, which run 5.

\begin{table}[!htbp]
\caption{Hyperparameter search spaces per method.}
\label{tab:hp-spaces}
\centering
\small\setlength{\tabcolsep}{4.5pt}
\begin{tabular}{llll}
\toprule
\textbf{Method} & \textbf{Hyperparameter} & \textbf{Search Space} & \textbf{Notes} \\
\midrule
VARLiNGAM        & none             & ---                              & Nothing tuned \\
VARLiNGAM\_imp   & none             & ---                              & Mean imputation pre-fit; not tuned \\
\midrule
PCMCI-ParCorr    & pc\_alpha         & \{0.001, 0.01, 0.05\}            &                              \\
\midrule
PCMCI-GPDC       & pc\_alpha         & \{0.001, 0.01, 0.05\}            & 5 trials per cell; $d=4$ only \\
\midrule
DYNOTEARS        & lambda\_w         & log-uniform $[10^{-3}, 1]$       & 5 trials per cell            \\
                 & lambda\_a         & log-uniform $[10^{-3}, 1]$       &                              \\
                 & w\_threshold      & fixed at 0.1                     & Not tuned                    \\
\midrule
CD-NOD           & alpha             & \{0.001, 0.01, 0.05\}            & Contemporaneous-only algorithm \\
\midrule
CUTS+            & lr                & log-uniform $[10^{-4}, 10^{-2}]$ &                              \\
                 & lambda\_sparsity  & log-uniform $[10^{-4}, 10^{-1}]$ &                              \\
                 & threshold         & uniform $[0.05, 0.5]$            & Hard threshold post-training \\
\midrule
GrangerF         & alpha             & \{0.001, 0.01, 0.05\}            &                              \\
\bottomrule
\end{tabular}
\end{table}

\section{Precision and recall decomposition}
\label{app:precision-recall}

Table~\ref{tab:prec-rec} decomposes skeleton F1 into precision and 
recall on the Linear DGP. Methods with high precision and lower recall 
exhibit conservative edge selection; methods with low precision and 
higher recall exhibit liberal thresholding. The decomposition appears 
in the main results table (Table~\ref{tab:main-discovery-obs}) under the 
TPR and FDR columns; this appendix retains a per-method view for 
readers familiar with the precision/recall framing.

\begin{table}[!htbp]
\caption{Precision and recall decomposition on Linear DGP. Mean 
over 90 runs per method.}
\label{tab:prec-rec}
\centering
\small
\begin{tabular}{lccc}
\toprule
 & Precision & Recall & F1 \\
\midrule
DYNOTEARS & \textbf{0.99} & \textbf{1.00} & \textbf{0.99} \\
PCMCI\_GPDC & 0.95 & 0.67 & 0.78 \\
VARLiNGAM & 0.73 & 0.80 & 0.76 \\
VARLiNGAM\_imp & 0.73 & 0.80 & 0.76 \\
PCMCI\_ParCorr & 0.88 & 0.65 & 0.74 \\
CUTS+ & 0.64 & 0.87 & 0.74 \\
GrangerF & 0.33 & 0.65 & 0.43 \\
CDNOD & 0.08 & 0.03 & 0.04 \\
\bottomrule
\end{tabular}

\end{table}

\section{Per-dimension performance breakdowns}
\label{app:dim-scaling}

Tables~\ref{tab:discovery-by-d} and~\ref{tab:cate-by-d} decompose 
discovery and CATE performance by graph dimension $d$. DYNOTEARS 
SHD remains essentially flat across $d \in \{4, 6, 8\}$, confirming 
that score-based continuous optimization scales gracefully on sparse 
graphs at this dimension range. PCMCI-ParCorr (0.41, 0.39, 0.56) and 
GrangerF (1.75, 1.78, 1.97) degrade at $d = 8$, while VARLiNGAM stays near 0.5. PCMCI-GPDC's GP-kernel cost excludes 
$d \in \{6, 8\}$ from our compute budget. On the CATE side, the Linear DGP gives a mixed 
picture. X-Learner's PEHE falls from 0.84 at $d = 4$ to 0.61 at $d = 8$, TARNet's goes 0.44, 0.59, 0.44, 
and SNet's and FlexTENet's rise. Higher dimension therefore does not uniformly degrade estimation, 
but we have not tested why.

\begin{table}[!htbp]
\centering
\caption{Performance breakdown by graph dimension on Linear DGP. Mean 
($\pm$ std) over 30 runs per cell (3 $\tau$ configurations $\times$ 10 
seeds), filtered to \texttt{status=ok}. PCMCI-GPDC at $d \in \{6,8\}$ 
was excluded by the compute-budget skip.}
\label{tab:discovery-by-d}
\small
\begin{tabular}{lllllll}
\toprule
 & \multicolumn{3}{c}{SHD} & \multicolumn{3}{c}{Runtime (s)} \\
d & 4 & 6 & 8 & 4 & 6 & 8 \\
\midrule
CDNOD & 1.00 & 1.03 & 1.06 & 0.01 & 0.01 & 0.02 \\
CUTS+ & 0.67 & 0.36 & 0.40 & 0.06 & 0.06 & 0.06 \\
DYNOTEARS & 0.00 & 0.02 & 0.03 & 0.15 & 0.41 & 0.59 \\
GrangerF & 1.75 & 1.78 & 1.97 & 0.02 & 0.04 & 0.07 \\
PCMCI\_GPDC & 0.38 & - & - & 41.78 & - & - \\
PCMCI\_ParCorr & 0.41 & 0.39 & 0.56 & 0.13 & 0.27 & 0.51 \\
VARLiNGAM & 0.55 & 0.47 & 0.53 & 0.15 & 0.27 & 0.44 \\
VARLiNGAM\_imp & 0.55 & 0.47 & 0.53 & 0.16 & 0.29 & 0.43 \\
\bottomrule
\end{tabular}

\end{table}

\begin{table}[!htbp]
\centering
\caption{CATE performance by dimension on Linear DGP. Mean ($\pm$ std) 
over 30 runs per cell (3 $\tau$ configurations $\times$ 10 seeds).}
\label{tab:cate-by-d}
\small
\begin{tabular}{lccc}
\toprule
 & \multicolumn{3}{c}{PEHE} \\
 & d=4 & d=6 & d=8 \\
\midrule
TARNet & \textbf{0.440 $\pm$ 0.222} & \textbf{0.594 $\pm$ 0.127} & \textbf{0.438 $\pm$ 0.097} \\
SNet & 0.458 $\pm$ 0.218 & 0.750 $\pm$ 0.143 & 0.814 $\pm$ 0.076 \\
FlexTENet & 0.478 $\pm$ 0.215 & 0.753 $\pm$ 0.132 & 0.756 $\pm$ 0.091 \\
X-Learner & 0.841 $\pm$ 0.155 & 0.820 $\pm$ 0.112 & 0.611 $\pm$ 0.060 \\
\bottomrule
\end{tabular}

\end{table}

\section{DYNOTEARS robustness to nonlinear mechanism variants}
\label{app:nonlin-monotonicity}

To probe the robustness boundary of DYNOTEARS' continuous DAG-learning
objective under nonlinear DGPs, we ran a targeted experiment varying only the
structural mechanism. Table~\ref{tab:nonlin-monotonicity} compares three
mechanisms on all nine reference graphs across the ten evaluation seeds
($n{=}90$ cells per mechanism), holding the graph, weights, noise realization,
and DYNOTEARS hyperparameters fixed; the hyperparameters are the values tuned
for each graph's Nonlinear cell. The mechanisms are Linear (identity),
Nonlinear ($\text{sign}(x)|x|^{0.5}$, strictly monotonic and saturating), and
a non-monotonic alternative $\sin(x)$.

As Table~\ref{tab:nonlin-monotonicity} shows, the strictly monotonic
saturating transform leaves DYNOTEARS' F1 at the Linear baseline (0.99 in
both cases). Substituting a non-monotonic but bounded transform produces
measurable but bounded degradation (F1 falls to 0.88 and mean SHD rises from
0.11 to 1.14), concentrated on a few graphs (mean SHD 3.0 on $d{=}6,\tau_{\max}{=}4$, 
2.3 on $d{=}8,\tau_{\max}{=}4$, and 2.2 on $d{=}4,\tau_{\max}{=}2$, against at most 1.1 
on the other six). We additionally evaluated a strictly monotonic but
amplifying transform $x \cdot |x|$. Recursive composition of this mechanism
inflates trajectories by orders of magnitude (peak $|x|$ above $10^4$ on four
of the six graphs with $d \geq 6$, up to $9 \times 10^7$ at $T{=}1000$), which
turns these cells into a numerical-conditioning test for any
linear-regression-based discovery method rather than a structural-recovery
test. DGP specifications that couple recursive amplification with
linear-regression-based discovery wrappers should therefore include a
stationarity guard; this is on the roadmap for future versions of the
library. The experiment is reproduced by
\texttt{scripts/run\_nonlinear\_mechanisms.py}.

\begin{table}[!htbp]
\centering
\small
\caption{DYNOTEARS performance under different recursive nonlinear
mechanisms. Mean ($\pm$ std) over 90 cells per mechanism (nine graphs
$\times$ ten seeds). Hyperparameters held fixed at each graph's
Nonlinear-cell tuning.}
\label{tab:nonlin-monotonicity}
\begin{tabular}{lccc}
\toprule
Mechanism & Property & SHD ($\downarrow$) & F1 ($\uparrow$) \\
\midrule
Linear ($x$) & identity & $0.10 \pm 0.30$ & $0.99 \pm 0.02$ \\
Nonlinear ($\text{sign}(x)|x|^{0.5}$) & monotonic, saturating & $0.11 \pm 0.38$ & $0.99 \pm 0.03$ \\
Non-monotonic ($\sin x$) & non-monotonic, bounded & $1.14 \pm 1.32$ & $0.88 \pm 0.15$ \\
\bottomrule
\end{tabular}

\end{table}

\section{Drift variant results}
\label{app:drift-table}

Table~\ref{tab:drift} reports performance on the Drift variant of 
the DGP, in which structural weights vary linearly with $t/T$. 
Performance is within 0.08 SHD of Linear for every 
method, consistent with the view that methods fitting the average structural 
relationship recover it whether the relationship is strictly 
time-invariant or smoothly time-varying.

\begin{table}[!htbp]
\centering
\caption{Causal discovery performance on Drift DGP. Mean ($\pm$ std) 
over 90 runs per cell.}
\label{tab:drift}
\small
\begin{tabular}{lccc}
\toprule
 & \multicolumn{3}{c}{Drift} \\
 & SHD & TPR & FDR \\
\midrule
DYNOTEARS & \textbf{0.04 $\pm$ 0.07} & \textbf{0.98 $\pm$ 0.05} & \textbf{0.01 $\pm$ 0.04} \\
PCMCI\_GPDC & 0.38 $\pm$ 0.17 & 0.67 $\pm$ 0.12 & 0.06 $\pm$ 0.12 \\
VARLiNGAM & 0.51 $\pm$ 0.36 & 0.81 $\pm$ 0.14 & 0.27 $\pm$ 0.17 \\
VARLiNGAM\_imp & 0.51 $\pm$ 0.36 & 0.81 $\pm$ 0.14 & 0.26 $\pm$ 0.17 \\
PCMCI\_ParCorr & 0.45 $\pm$ 0.16 & 0.65 $\pm$ 0.07 & 0.12 $\pm$ 0.13 \\
CUTS+ & 0.48 $\pm$ 0.22 & 0.81 $\pm$ 0.33 & 0.33 $\pm$ 0.26 \\
GrangerF & 1.84 $\pm$ 0.57 & 0.65 $\pm$ 0.07 & 0.67 $\pm$ 0.09 \\
CDNOD & 1.04 $\pm$ 0.11 & 0.03 $\pm$ 0.06 & 0.92 $\pm$ 0.18 \\
\bottomrule
\end{tabular}

\end{table}

\section{Failed run error distribution}
\label{app:failures}

Across the 6{,}480 (method, graph, variant, seed) combinations 
attempted, 4{,}980 succeeded, 960 were skipped because the method 
declared \texttt{supports\_missing=False} on a missingness variant, 
540 were skipped on PCMCI-GPDC at $d > 4$ by the compute-budget 
guard, and none raised an exception. Table~\ref{tab:failures-summary} 
aggregates these counts by method.

\begin{table}[!htbp]
\centering
\caption{Aggregate failure mode counts by method, summed across all 
nine DGP variants.}
\label{tab:failures-summary}
\small\setlength{\tabcolsep}{4.5pt}
\begin{tabular}{lrrrr}
\toprule
 & Completed & Skipped (no missing support) & Skipped (compute budget) & Errored \\
\midrule
CDNOD & 630 & 180 & 0 & 0 \\
CUTS+ & 810 & 0 & 0 & 0 \\
DYNOTEARS & 630 & 180 & 0 & 0 \\
GrangerF & 630 & 180 & 0 & 0 \\
PCMCI\_GPDC & 210 & 60 & 540 & 0 \\
PCMCI\_ParCorr & 630 & 180 & 0 & 0 \\
VARLiNGAM & 630 & 180 & 0 & 0 \\
VARLiNGAM\_imp & 810 & 0 & 0 & 0 \\
\bottomrule
\end{tabular}

\end{table}

The first run of this benchmark, before the wrapper fix, had 171 errors, all in 
VARLiNGAM\_imp (4{,}809 completed rows in total). They were not specific to missing data. They
occurred on complete data as well (15 of 90 Linear cells), and they traced to the
wrapper defect described below rather than to the imputed covariance matrices.
Table~\ref{tab:varlingam-errors} shows the per-variant counts after the fix. 
CUTS+, the only other method that supports missing values, completed 
all 90 cells under both MCAR and MNAR.

\begin{table}[!htbp]
\centering
\caption{VARLiNGAM\_imp completion vs. error counts by DGP variant, after the wrapper fix. 
Each variant has 90 attempted runs (9 graphs $\times$ 10 seeds).}
\label{tab:varlingam-errors}
\small
\begin{tabular}{lrrr}
\toprule
status & Completed & Errored & Total attempted \\
\midrule
Linear & 90 & 0 & 90 \\
Nonlinear & 90 & 0 & 90 \\
Drift & 90 & 0 & 90 \\
Trend & 90 & 0 & 90 \\
Seasonal & 90 & 0 & 90 \\
Trend+Seasonal & 90 & 0 & 90 \\
Nonlinear+Trend+Seasonal & 90 & 0 & 90 \\
MCAR & 90 & 0 & 90 \\
MNAR & 90 & 0 & 90 \\
\bottomrule
\end{tabular}

\end{table}

\paragraph{Root cause and correction.} The VARLiNGAM\_imp failures in our first run
were traced after submission to a wrapper defect, not to the
method: \texttt{lingam.VARLiNGAM} selects the VAR lag order by BIC, so its
adjacency array can be shorter than $\tau_{\max}{+}1$, and our wrapper indexed
past its end. Because the failing cells were silently dropped from means, and
they were exactly the cells where BIC chose a short lag order, the VARLiNGAM\_imp scores
we first obtained were survivorship-biased upward (nonlinear variant:
oriented $F_1$ $0.85$ to a corrected $0.76$ under the audit protocol). The
wrapper is fixed, and the regeneration reported in this paper completes all nine
variants with zero failures. On standardized data, and on raw data from the four stationary
variants, VARLiNGAM and VARLiNGAM\_imp then agree in every cell, as they should,
since imputation is a no-op on complete data. On raw data from the five trend,
seasonal, and drift variants they differ in 125 of 450 cells, because the
VARLiNGAM wrapper standardizes its input before fitting while the imputation
wrapper fits on the raw scale.

\paragraph{CUTS+ empty graphs.} In 3 of the 81 CUTS+ cells (d4\_tau4 Linear, d4\_tau4 Drift, and d6\_tau3 Seasonal) 
the tuned setting predicts no edges on any of the ten seeds, which scores F$_1 = 0$ for the cell and 
lowers the Linear and Drift means in the main tables. The tuning is deterministic, but which cells collapse 
depends on the tuning trajectory: the first run collapsed in four other cells.

\paragraph{Determinism of the regeneration.} The reference benchmark is
seeded end to end. The Optuna TPE sampler is seeded from the tuning-cache key,
which covers the method, cell, tuning seed, trial budget, and search-space
version, and the CATE estimators receive the evaluation seed. We reran selected
cells (four discovery cells with ten seeds each, and two CATE estimators with
three seeds) in separate processes and obtained identical rows, with identical
tuning caches. We have not tested other machines or package versions; the
environment (Python 3.12.13, package versions) is recorded in
\texttt{environment.json} in the repository.

\section{Scalability beyond the reference benchmark}
\label{app:scalability}

The reference benchmark holds $d \in \{4,6,8\}$ and $T{=}1000$ for tractability.
Table~\ref{tab:scalability} extends the sweep to $d \in \{4,8,16,32\}$ and
$T \in \{1000, 5000, 10000\}$ at $\tau_{\max}{=}3$, on Erd\H{o}s-R\'enyi window
graphs with $\approx d$ edges, three seeds per cell. Only methods whose
implementation is vendored here or depends on core requirements are included, so
the sweep reproduces without the optional extras. The graph is drawn once per
$d$ and held fixed, so variation across seeds reflects the noise realization
only, consistent with the protocol in Section~\ref{sec:experiments}.

\begin{table}[!htbp]
\centering
\caption{Scalability sweep at $\tau_{\max}{=}3$, three seeds per cell (one seed
at $d{=}32, T{=}10000$). ``std.'' is the variance-standardized condition of
Section~\ref{sec:varsortability}. Runtime is single-core wall clock for the raw
condition.}
\label{tab:scalability}
\small
\begin{tabular}{ccccccc}
\toprule
& & \multicolumn{2}{c}{DYNOTEARS F1} & \multicolumn{2}{c}{\textsc{SortnRegress} F1} & DYNOTEARS \\
\cmidrule(lr){3-4}\cmidrule(lr){5-6}
$d$ & $T$ & raw & std. & raw & std. & runtime (s) \\
\midrule
4 & 1000 & 1.00 & 1.00 & 0.70 & 0.85 & 0.2 \\
4 & 5000 & 1.00 & 1.00 & 0.70 & 0.55 & 0.3 \\
4 & 10000 & 1.00 & 1.00 & 0.55 & 0.85 & 0.4 \\
8 & 1000 & 1.00 & 0.82 & 0.70 & 0.75 & 0.3 \\
8 & 5000 & 1.00 & 0.82 & 0.74 & 0.71 & 0.6 \\
8 & 10000 & 1.00 & 0.82 & 0.74 & 0.71 & 1.3 \\
16 & 1000 & 1.00 & 0.81 & 0.94 & 0.83 & 1.0 \\
16 & 5000 & 1.00 & 0.81 & 0.98 & 0.73 & 1.9 \\
16 & 10000 & 1.00 & 0.81 & 0.98 & 0.71 & 3.3 \\
32 & 1000 & 1.00 & 0.91 & 0.94 & 0.76 & 5.6 \\
32 & 5000 & 1.00 & 0.92 & 0.92 & 0.73 & 7.0 \\
32 & 10000 & 1.00 & 0.94 & 0.98 & 0.67 & 13.4 \\
\bottomrule
\end{tabular}

\end{table}

Two observations. First, DYNOTEARS attains $F_1 = 1.00$ on raw data at
\emph{every} cell, including $d{=}32, T{=}10000$. This is not evidence of
favorable scaling. \textsc{SortnRegress} also rises with dimension on raw data,
which indicates that the marginal-variance shortcut becomes \emph{more}
informative as graphs grow, because longer structural paths accumulate more
variance separation. Scaling the benchmark up without standardizing would
therefore have made it easier, not harder. The standardized condition is the one
in which the scaling question is meaningful, and the full grid below is read
that way.

\paragraph{Full grid.} Table~\ref{tab:scal-f1} and Table~\ref{tab:scal-runtime}
extend the sweep to all nine methods, $d \in \{4,8,16,32\}$,
$T \in \{1000,5000\}$, and $\tau_{\max} \in \{3,6,12\}$ in both conditions,
three seeds per cell: 1{,}296 method-cells in total, every one of which
completed with status \texttt{ok} (no timeouts against a 900\,s per-fit budget;
slowest single fit 193\,s). The sweep runs every method at library defaults,
unlike the tuned reference benchmark, and the random-graph family here is
sparser and lower-varsortability than the reference graphs (contemporaneous edge
fraction $0.25$ to $0.31$ against $0.38$ to $0.50$; per-cell $v$ spans $0.0$ to
$1.0$ and is recorded in the released CSVs), so absolute $F_1$ is not comparable
across the two tables and cross-method comparison is not meaningful here. We
therefore report the grid as \emph{within-method} scaling plus runtime.

\begin{table}[!htbp]
\centering
\caption{Standardized-condition $F_1$ of the leading method across the full
grid, mean over $T$ and three seeds. The minimum over all twelve
$(d, \tau_{\max})$ cells is $0.91$: once the variance shortcut is removed,
accuracy does not degrade with dimension, trajectory length, or lag order.}
\label{tab:scal-f1}
\small
\begin{tabular}{lcccc}
\toprule
DYNOTEARS $F_1$ (std.) & $d{=}4$ & $d{=}8$ & $d{=}16$ & $d{=}32$ \\
\midrule
$\tau_{\max}{=}3$ & 1.00 & 0.94 & 0.92 & 0.98 \\
$\tau_{\max}{=}6$ & 1.00 & 0.96 & 0.95 & 0.91 \\
$\tau_{\max}{=}12$ & 1.00 & 1.00 & 0.97 & 0.98 \\
\bottomrule
\end{tabular}

\end{table}

\begin{table}[!htbp]
\centering
\caption{Mean seconds per fit at the largest dimension ($d{=}32$, raw
condition; standardized is comparable). The binding runtime constraint shifts
with lag order: VARLiNGAM is flat in $\tau_{\max}$ and binds at
$\tau_{\max}{=}3$, PCMCI-ParCorr grows roughly linearly in the window and binds
at $\tau_{\max}{=}6$, and GrangerF grows fastest and binds at
$\tau_{\max}{=}12$.}
\label{tab:scal-runtime}
\small
\begin{tabular}{lccc}
\toprule
Seconds per fit, $d{=}32$ & $\tau_{\max}{=}3$ & $\tau_{\max}{=}6$ & $\tau_{\max}{=}12$ \\
\midrule
DYNOTEARS & 5 & 9 & 16 \\
VARLiNGAM & 25 & 24 & 21 \\
PCMCI-ParCorr & 22 & 47 & 118 \\
GrangerF & 14 & 45 & 187 \\
\bottomrule
\end{tabular}

\end{table}

Two default-configuration rows require a caption rather than a comparison. At
library defaults CUTS+ returns an empty graph on this family (the wrapper's
edge threshold zeroes every coefficient at the default learning rate), which
forces $F_1 = 0$ arithmetically; at the hyperparameters the reference benchmark
tuned for it, spot checks on the same cells give $0.82$ at $d{=}8$ and $0.86$
at $d{=}16$. Its default-configuration row is an artifact of our wrapper's
defaults, not a property of the method or of scaling. CDNOD, by contrast, does
predict edges at defaults and its low score reflects a genuine method-DGP
mismatch on a sparse linear family that favors SVAR learners.

\paragraph{Varsortability as an instrument.} Because $v$ varies naturally
across this random-graph family, the sweep doubles as a check on the measure
itself. Over the 72 raw/standardized cell pairs, the $F_1$ that
\textsc{SortnRegress} loses to standardization is monotone in $v$: $-0.235$
where $v = 0.0$ (variance ordering anti-correlates with causal order, so
removing it helps), $+0.016$ near $v = 0.5$, and $+0.234$ for $v \in (0.9, 1]$
(Pearson $r = 0.77$). \textsc{RandomRegress} shows exactly zero change in every
cell, and the scale-invariant methods are identical throughout. We note the
scope of this check: $v$ measures whether variance ordering matches causal
ordering, which is precisely what \textsc{SortnRegress} exploits, so the tight
tracking validates the measure and confirms that standardization removes
exactly what it targets, rather than independently proving that real methods
ride the shortcut; that evidence is the reference-benchmark contrast of
Section~\ref{sec:varsortability}, and consistent with being a method rather
than a pure exploiter, DYNOTEARS tracks $v$ far more weakly ($r = 0.40$).

\section{An external equation family: TimeGraph C1}
\label{app:timegraph}

The limitations section notes that a synthetic generator's mechanism class
tends to resemble the model forms prevailing when the generator is written.
To test the reference results against an independently authored family, we
evaluated the same methods on data produced by the C1 generator of
TimeGraph \citep{Ferdous2025TimeGraph}, a benchmark published before
\libname{} and designed without reference to it. For transparency we note
that TimeGraph is earlier work by the present authors; its equations
nonetheless constitute an external family in the relevant sense, since
neither the mechanisms nor their parameters were chosen with \libname{}'s
reference benchmark in mind.

\paragraph{Setup.} The generator class was imported unmodified from the
published TimeGraph repository, and ground truth was parsed from its
equation strings and verified against the known structure of each
configuration. C1 composes per-node structural terms drawn from sines,
cosines, squares, and cubes with a per-channel linear trend, a two-harmonic
seasonal component with period 12, and Gaussian noise. We ran all nine
published configurations ($d \in \{4,6,8\}$, maximum lag in $\{2,3,4\}$) at
$T{=}1000$ with two seeds per cell, in raw and standardized conditions,
with per-cell varsortability recorded, under the same oriented window-graph
$F_1$ as the reference benchmark. Three variants separate the exogenous
components. The \emph{full} variant uses TimeGraph defaults (trend strength
$0.01$, seasonal strength $0.5$); \emph{seasonal} sets the trend to zero;
\emph{mechanism} sets both to zero, a small-amplitude regime in which the
sine mechanisms are effectively linear and the even-function edges carry
little signal. DYNOTEARS received an oracle-tuned advantage, a
$3{\times}2$ regularization grid selected on a held-out seed per
configuration and condition in the full and seasonal variants, while every
other method ran at library defaults. All 108 method-cells completed with
zero failed fits.

\begin{table}[!htbp]
\centering
\caption{Mean oriented $F_1$ on the TimeGraph C1 family (nine
configurations, two seeds), with contemporaneous varsortability $v$ per
variant. $^{\dagger}$DYNOTEARS is oracle-tuned on a held-out seed in the
full and seasonal variants and runs at library defaults in the mechanism
variant; all other methods run at library defaults throughout. For
reference, tuned DYNOTEARS reaches $0.99$ (raw) on the linear reference
family of Section~\ref{sec:varsortability}.}
\label{tab:tg-c1}
\small
\begin{tabular}{lcccccc}
\toprule
 & \multicolumn{2}{c}{full ($v{=}1.00$)} & \multicolumn{2}{c}{seasonal ($v{=}0.56$)} & \multicolumn{2}{c}{mechanism ($v{=}0.25$)} \\
\cmidrule(lr){2-3}\cmidrule(lr){4-5}\cmidrule(lr){6-7}
Method & raw & std. & raw & std. & raw & std. \\
\midrule
DYNOTEARS$^{\dagger}$ & 0.13 & 0.19 & 0.21 & 0.24 & 0.00 & 0.49 \\
\textsc{SortnRegress} & 0.14 & 0.11 & 0.11 & 0.09 & 0.51 & 0.48 \\
\textsc{RandomRegress} & 0.11 & 0.11 & 0.10 & 0.10 & 0.44 & 0.44 \\
VARLiNGAM & 0.11 & 0.11 & 0.09 & 0.09 & 0.44 & 0.44 \\
GrangerF & 0.06 & 0.06 & 0.09 & 0.09 & 0.30 & 0.30 \\
PCMCI-ParCorr & 0.07 & 0.07 & 0.08 & 0.08 & 0.31 & 0.31 \\
\bottomrule
\end{tabular}

\end{table}

\paragraph{Family dependence.} On the seasonal variant, the fair isolation
of the equations, oracle-tuned DYNOTEARS reaches $0.21$ raw against $0.99$
on the reference linear family, and every untuned method sits at or near
the \textsc{RandomRegress} floor. The reason is mechanistic rather than a
matter of tuning. C1's edges are sines, cosines, squares, and cubes, and
the even functions among them carry no linear signal, so linear-SVAR and
partial-correlation machinery cannot represent most of the graph.
Benchmark scores are therefore properties of a (method, family) pair, and
the near-ceiling reference results do not transfer to an equation family
the generator's authors wrote for a different purpose.

\paragraph{Varsortability as a per-cell diagnostic.} The three variants
exercise the recorded $v$ metadata across its range. The seasonal variant
measures $v = 0.56$, essentially uninformative, and standardization
changes little, confirming that this family carries no variance shortcut
and that its difficulty lies in the mechanisms. The full variant is
perfectly varsortable ($v = 1.00$) because the per-channel trends grow
with the variable index, yet \textsc{SortnRegress} scores only $0.14$;
under this degree of non-stationarity the ordering information exists but
the regression step cannot exploit it, so high $v$ is necessary but not
sufficient for the shortcut to operate.

\paragraph{A second default-configuration artifact.} In the
small-amplitude mechanism variant, DYNOTEARS at library defaults returns
an empty graph on raw data ($F_1 = 0.00$ against $0.49$ standardized)
because the default penalty dominates coefficients fit to signals with
tiny variance. This is the same failure mode documented for CUTS+ in
Appendix~\ref{app:scalability}, now in a second method, and it reinforces
the caption practice adopted there, that default-configuration rows
describe configurations, not methods. The scale-invariant methods are
identical across conditions in every variant, as they must be.

\paragraph{Scope.} Two seeds per cell; absolute $F_1$ is not comparable
across families, since graphs and edge counts differ; the reproduction
script ships with the code release.

\section{Per-method computational cost}
\label{app:compute}

The reference benchmark was regenerated on an Apple M4 Mac mini (10 cores, 24 GB) 
in about 2.6 hours, with four worker processes for discovery and for CATE. 
Per-fit runtimes are measured under that parallel load, so we compare them 
across methods and not across machines. 
Per-method runtime broken down by dimension on the Linear DGP appears 
in Table~\ref{tab:discovery-by-d}. PCMCI-GPDC dominates absolute 
runtime due to GP kernel evaluations, motivating the $d=4$ restriction 
in the main benchmark. DYNOTEARS scales modestly with $d$; 
PCMCI-ParCorr and VARLiNGAM scale near-linearly. Memory profiling is 
omitted from the reference benchmark in this version.

\section{Spatio-temporal extension details}
\label{app:spatial}

Beyond the within-trajectory framework above, \libname{} provides a spatio-temporal module that lifts the structural equations to grids 
of locations indexed by $(t, \ell, i)$ for spatial position $\ell$ 
and within-location variable $i$. Locations interact through a 
configurable spatial kernel $K \in \mathbb{R}^{L \times L}$, with 
$K[\ell', \ell]$ quantifying the influence of location $\ell'$ on 
$\ell$. Built-in topologies include 2D and 3D grids and random 
geometric graphs; built-in kernels include Gaussian, power-law, 
and step kernels, with a generic adjacency-based kernel for 
user-supplied spatial structures. The cached-noise rollout 
mechanism extends naturally to this setting, yielding exact 
counterfactual twins for spatial interventions (e.g., a hard 
intervention $do(X_{t,\ell^*,i} = c)$ at a single location $\ell^*$ 
propagating to neighboring locations through $K$). Identifiability 
guarantees follow analogous classification: the spatial kernel is 
identifiable under the standard separability assumption that 
$K$ is constant across time and that within-location and across-location 
mechanisms factor cleanly. We exclude the spatio-temporal regime 
from the reference benchmark in Section~\ref{sec:experiments} 
because no widely-adopted spatio-temporal causal discovery method 
has emerged in the time-series literature; the module's role is 
to provide a substrate for follow-up methodological work, not to 
serve as a current evaluation target.

\section{Datasheet for \libname{}}
\label{app:datasheet}

We follow the structure proposed by~\citet{gebru2021datasheets}.

\subsection*{Motivation}
\textbf{For what purpose was the dataset created?} \libname{} was created 
to provide a configurable benchmark with documented identifiability status for evaluating 
time-series causal discovery and dynamic treatment-effect estimation 
methods, addressing the empirical gap between synthetic benchmark 
performance and real-world deployment robustness.

\textbf{Who created the dataset?} Muhammad Hasan Ferdous and Md Osman Gani,
Department of Information Systems, University of Maryland, Baltimore County.
Funding is disclosed in the Acknowledgments and Disclosure of Funding.

\subsection*{Composition}
\textbf{What do the instances represent?} Each instance is a multivariate 
time-series trajectory $\{X_{t,i}\}_{t=1,i=1}^{T,d}$ paired with its 
ground-truth contemporaneous and lagged adjacency matrices, the cached 
noise realization $\mathbf{N}$, the observational mask $R$ when applicable, 
and (for CATE cells) the counterfactual twin trajectory under a specified 
intervention.

\textbf{How many instances are there?} The reference benchmark contains 
$9 \times 9 \times 10 = 810$ discovery instances per method (total 6,480 
discovery rows before status filtering) and $9 \times 2 \times 10 = 180$ 
CATE instances per method (total 720). The framework generates additional 
instances on demand under any user-supplied configuration.
\paragraph{Spatio-temporal instances.}
The framework also supports spatio-temporal grid instances of shape 
$(T, L, d)$ for $L$ locations on a configurable topology, paired with 
the spatial kernel matrix and cached noise. These are not part of the 
reference benchmark in this paper but are accessible through the 
\texttt{dtcd.spatial} module.

\textbf{Are there sensitive attributes or PII?} No. All data are synthetic.

\subsection*{Collection process}
\textbf{How was the data acquired?} Generated by the \libname{} simulator 
according to the documented DGP (Section~\ref{sec:framework}).

\subsection*{Preprocessing / cleaning / labeling}
Ground-truth adjacency matrices are produced jointly with the data. No 
post-hoc cleaning is performed. Splits are defined at the seed level. Seeds
$\{42,\ldots,51\}$ are used for evaluation, and hyperparameters are tuned on
a separate trajectory generated at seed $41$, which is never evaluated.

\subsection*{Uses}
\textbf{Has the dataset been used already?} For benchmarking the methods 
in Section~\ref{sec:experiments}.

\textbf{Are there tasks for which the dataset should not be used?} 
Strong performance on \libname{} does not guarantee strong real-world 
deployment performance. \libname{} should not be used as the sole evidence 
for deployment decisions in safety-critical applications.

\subsection*{Distribution and maintenance}
\textbf{License.} Library: Apache-2.0. Datasets: CC-BY-4.0.

\textbf{Maintenance.} Issues and feature requests are tracked at
\url{https://github.com/hferdous/Composable_Causality/issues}. Versioned
releases follow semantic versioning.

\section*{NeurIPS Paper Checklist}

\begin{enumerate}

\item {\bf Claims}
    \item[] Question: Do the main claims made in the abstract and introduction accurately reflect the paper's contributions and scope?
    \item[] Answer: \answerYes{}
    \item[] Justification: The abstract and Introduction (Section~\ref{sec:intro}) clearly enumerate \libname{}'s contributions: a composable framework, exact counterfactual twin generation, identifiability documentation, the reference benchmark across 81 configurations, and open-source release. Section~\ref{sec:experiments} demonstrates these claims empirically, and Section~\ref{sec:discussion} acknowledges where conclusions are scoped.

\item {\bf Limitations}
    \item[] Question: Does the paper discuss the limitations of the work performed by the authors?
    \item[] Answer: \answerYes{} 
    \item[] Justification: Section~6.1 explicitly discusses five limitations: synthetic-to-real gap, identifiability boundaries (unestablished cases are named explicitly, and machine-readable tags exist only in the spatio-temporal module), computational scope (including the PCMCI-GPDC $d=4$ restriction), weight-realization sensitivity (the protocol fixes weights per graph), and the counterfactual sufficiency assumption.

\item {\bf Theory assumptions and proofs}
    \item[] Question: For each theoretical result, does the paper provide the full set of assumptions and a complete (and correct) proof?
    \item[] Answer: \answerYes{}
    \item[] Justification: Propositions 1 and 2 (Section~3.2) state identifiability claims for Linear Gaussian VAR and VAR-LiNGAM with their assumptions explicit (causal sufficiency, Gaussianity or non-Gaussianity, acyclicity at each time slice). The recursive nonlinear power-transform mechanism is explicitly marked as identifiability-unknown and presented without an identifiability claim.

    \item {\bf Experimental result reproducibility}
    \item[] Question: Does the paper fully disclose all the information needed to reproduce the main experimental results of the paper to the extent that it affects the main claims and/or conclusions of the paper (regardless of whether the code and data are provided or not)?
    \item[] Answer: \answerYes{}
    \item[]Justification: Section~\ref{sec:experiments} fully specifies the benchmark protocol: nine graph templates, nine discovery DGP variants, ten evaluation seeds (42--51), the structural weight seed derivation rule (per graph, fixed across noise seeds), all hyperparameter search spaces (Appendix~\ref{app:hyperparameters}), and the structured failure-reporting protocol. Source code, configuration files, and seed lists are released in the public repository, where \texttt{bash reproduce.sh} rebuilds every table and figure from the released results and \texttt{bash reproduce.sh --full} reruns every experiment.

\item {\bf Open access to data and code}
    \item[] Question: Does the paper provide open access to the data and code, with sufficient instructions to faithfully reproduce the main experimental results, as described in supplemental material?
    \item[] Answer: \answerYes{}
    \item[] Justification: The library, generation scripts, baseline wrappers, and benchmark data are released under Apache-2.0 (code) and CC-BY-4.0 (datasets) licenses in the public repository linked from the abstract (\url{https://github.com/hferdous/Composable_Causality}), with documentation and Croissant metadata.

\item {\bf Experimental setting/details}
    \item[] Question: Does the paper specify all the training and test details (e.g., data splits, hyperparameters, how they were chosen, type of optimizer) necessary to understand the results?
    \item[] Answer: \answerYes{}
    \item[] Justification: Section~\ref{sec:experiments} specifies the optimizer (Optuna TPE), trial budget (30 per cell, reduced to 5 for PCMCI-GPDC and DYNOTEARS), a held-out tuning seed (41) never used for evaluation, tuning objective (F1 against ground truth), evaluation seeds (42--51), and the weight-fixing protocol. Appendix~\ref{app:hyperparameters} provides per-method hyperparameter search spaces.

\item {\bf Experiment statistical significance}
    \item[] Question: Does the paper report error bars suitably and correctly defined or other appropriate information about the statistical significance of the experiments?
    \item[] Answer:  \answerYes{}
    \item[] Justification: All tables report mean $\pm$ standard deviation across the ten evaluation seeds (42--51) within each cell. Standard deviations capture noise-realization variance under fixed weights, as documented in Section~\ref{sec:experiments}. The structured failure-reporting protocol distinguishes successful runs from skipped or errored runs (Tables~\ref{tab:failures-summary} and~\ref{tab:varlingam-errors}).
    

\item {\bf Experiments compute resources}
    \item[] Question: For each experiment, does the paper provide sufficient information on the computer resources (type of compute workers, memory, time of execution) needed to reproduce the experiments?
    \item[] Answer: \answerYes{}
    \item[] Justification: The reference benchmark (6,480 discovery cells + 720 CATE cells with hyperparameter tuning) ran on a single Mac Mini (Apple M4, 24GB RAM, no GPU) in approximately 4 hours wall-clock time. Per-method runtime is reported in Tables~\ref{tab:cate-main} (CATE) and~\ref{tab:discovery-by-d} (discovery).
    
\item {\bf Code of ethics}
    \item[] Question: Does the research conducted in the paper conform, in every respect, with the NeurIPS Code of Ethics \url{https://neurips.cc/public/EthicsGuidelines}?
    \item[] Answer: \answerYes{}
    \item[] Justification: \libname{} is a synthetic data generation library producing benchmark data with no PII, no human subjects, and no real-world decisions or deployments. The research conforms with the NeurIPS Code of Ethics in all respects.

\item {\bf Broader impacts}
    \item[] Question: Does the paper discuss both potential positive societal impacts and negative societal impacts of the work performed?
    \item[] Answer: \answerYes{}
    \item[] Justification: Section~6.2 discusses positive impacts (rigorous methodology floor, identifiability documentation reducing miscredited methods) and the absence of foreseeable misuse pathways specific to a synthetic benchmark.

\item {\bf Safeguards}
    \item[] Question: Does the paper describe safeguards that have been put in place for responsible release of data or models that have a high risk for misuse (e.g., pre-trained language models, image generators, or scraped datasets)?
    \item[] Answer: \answerNA{}
    \item[] Justification: \libname{} releases synthetic data generation code and benchmark numbers, not models with high misuse risk. There is no scraped data, no pretrained generative model, and no individually identifiable content.

\item {\bf Licenses for existing assets}
    \item[] Question: Are the creators or original owners of assets (e.g., code, data, models), used in the paper, properly credited and are the license and terms of use explicitly mentioned and properly respected?
    \item[] Answer: \answerYes{}
    \item[] Justification: All baseline implementations are credited via citations in Section~2 and used under their original licenses. The DYNOTEARS solver vendored from CausalNex (Apache-2.0) and the TimeGraph C1 generator (MIT) keep their license notices in the repository. Tigramite (GPL-3.0), causal-learn (MIT), lingam (MIT), CATENets (BSD-3-Clause), and EconML (MIT) are optional dependencies installed from their own distributions and are not redistributed.

\item {\bf New assets}
    \item[] Question: Are new assets introduced in the paper well documented and is the documentation provided alongside the assets?
    \item[] Answer: \answerYes{}
    \item[] Justification: \libname{} (the new asset) is documented via this paper, the Datasheet for Datasets in Appendix~\ref{app:datasheet}, Croissant metadata accompanying the artifact, and per-method docstrings in the released codebase.

\item {\bf Crowdsourcing and research with human subjects}
    \item[] Question: For crowdsourcing experiments and research with human subjects, does the paper include the full text of instructions given to participants and screenshots, if applicable, as well as details about compensation (if any)? 
    \item[] Answer: \answerNA{}
    \item[] Justification: This paper does not involve crowdsourcing or human subjects research. All data are synthetic.

\item {\bf Institutional review board (IRB) approvals or equivalent for research with human subjects}
    \item[] Question: Does the paper describe potential risks incurred by study participants, whether such risks were disclosed to the subjects, and whether Institutional Review Board (IRB) approvals (or an equivalent approval/review based on the requirements of your country or institution) were obtained?
    \item[] Answer: \answerNA{}
    \item[] Justification: This paper does not involve human subjects research; no IRB approval is required.

\item {\bf Declaration of LLM usage}
    \item[] Question: Does the paper describe the usage of LLMs if it is an important, original, or non-standard component of the core methods in this research? Note that if the LLM is used only for writing, editing, or formatting purposes and does \emph{not} impact the core methodology, scientific rigor, or originality of the research, declaration is not required.
    \item[] Answer: \answerYes{}
    \item[] Justification: LLMs were used for editorial assistance (text refinement, code refactoring suggestions, literature search verification) and as coding assistants that implemented and ran experiment scripts under the authors' direction, as disclosed in the submission form. LLMs were not used as a methodological component of the data generation framework, the discovery or estimation algorithms, or the experimental analysis.

\end{enumerate}

\end{document}